\documentclass[10pt,journal,draftcls,letterpaper,onecolumn]{IEEEtran}
\usepackage{times}
\usepackage{epsfig}
\usepackage{epstopdf}
\usepackage{graphicx}
\usepackage{amsmath}
\usepackage{amssymb}
\usepackage{amsthm}
\usepackage{cite}
\usepackage{booktabs}
\usepackage{algorithmic}
\usepackage{algorithm}
\usepackage{tabularx}
\usepackage{longtable}
\usepackage{bm}
\usepackage{color}
\usepackage[pagebackref=true,breaklinks=true,letterpaper=true,colorlinks,citecolor=blue,linkcolor=blue,bookmarks=false]{hyperref}

\usepackage{url}
\usepackage{caption}
\newcommand{\ignore}[1]{}

\usepackage{graphicx}
\usepackage{amsmath,amssymb} 
\usepackage{color}

\ifCLASSOPTIONcompsoc
\else
\fi

\begin{document}
%
\title{Robust Visual Tracking via Convolutional Networks}

\author{{\quad\\}Kaihua~Zhang, Qingshan~Liu, 
       Yi~Wu, and Ming-Hsuan~Yang
\IEEEcompsocitemizethanks{
\IEEEcompsocthanksitem Kaihua~Zhang, Qingshan~Liu and Yi~Wu are with Jiangsu Key Laboratory of Big Data Analysis Technology (B-DAT), Nanjing University of Information Science and Technology.
E-mail: \{cskhzhang, qsliu, ywu\}@nuist.edu.cn.
\IEEEcompsocthanksitem Ming-Hsuan Yang is with Electrical Engineering and Computer Science, University of California, Merced, CA, 95344. E-mail: mhyang@ucmerced.edu.}
\thanks{}}

\markboth{Submitted}%
{Shell \MakeLowercase{\textit{et al.}}: Bare Demo of IEEEtran.cls for Computer Society Journals}
\IEEEcompsoctitleabstractindextext{%
\begin{abstract}
Deep networks have been successfully applied to visual tracking by
learning a generic representation offline from numerous training
images.
However the offline training is time-consuming and the learned generic
representation may be less discriminative for tracking specific
objects.
In this paper we present that, even without offline training with a large amount of auxiliary data, simple two-layer
convolutional networks can  be powerful enough to develop a robust
representation for visual tracking.
In the first frame, we employ the $k$-means algorithm to extract a set of normalized patches
from the target region as fixed filters, which integrate a series of adaptive contextual filters surrounding the target to define a set of feature maps
in the subsequent frames.
These maps measure similarities between each filter and the useful
local intensity patterns across the target, thereby encoding its local
structural information.
Furthermore, all the maps form together a global representation, which
is built on mid-level features, thereby remaining close to image-level information, and hence the inner geometric layout of the target is also
well preserved.
A simple soft shrinkage method with an adaptive threshold is employed to de-noise the global representation, resulting in a robust sparse representation. The representation is updated via a simple and effective online strategy, allowing it to robustly adapt to target appearance
variations.
Our convolution networks have surprisingly lightweight structure, yet perform
favorably against several state-of-the-art methods on the CVPR2013 tracking
benchmark dataset with 50 challenging videos.
\end{abstract}

}

\maketitle

\IEEEdisplaynotcompsoctitleabstractindextext

%
\IEEEpeerreviewmaketitle

\section{Introduction}
Visual tracking is a fundamental problem in computer vision with a
wide range of  applications. Although much progress has been made in
recent years~\cite{Avidan_PAMI_2004, Kwon_CVPR_2010,
  Kalal_CVPR_2010,Hare_ICCV_2011, Zhang_ECCV_2012, wang2013learning},
it remains a challenging task due to many factors such as illumination
changes, partial occlusion, deformation, as well as viewpoint
variation (refer to~\cite{wu2013online}).
To address these challenges for robust tracking, recent
state-of-the-art
approaches~\cite{Kalal_CVPR_2010,Kwon_CVPR_2010,jia2012visual,dinh2011context,Hare_ICCV_2011,zhong2012robust}
focus on exploiting robust representations with hand-crafted features
(e.g., local binary patterns~\cite{Kalal_CVPR_2010}, Haar-like  features~\cite{Hare_ICCV_2011},
histograms~\cite{jia2012visual,zhong2012robust}, HOG descriptors~\cite{henriques2015high}, and covariance descriptors~\cite{gao2014transfer}).
However, these hand-crafted features are not tailored for all generic
objects, and hence require some sophisticated learning
techniques to improve  their representative capabilities.

Deep networks can directly learn features from raw data without
resorting to manual tweaking, which have gained much attention with
state-of-the-art results in some complicated tasks, such as
image classification~\cite{krizhevsky2012imagenet},
object recognition~\cite{donahue2014decaf}, detection and
segmentation~\cite{girshick2013rich}.
However, considerably less attention has been made to apply deep
networks for visual tracking.
The main  reason may be that there exists scarce amount of data to
train deep networks in visual tracking because only the target state
(i.e., position and size) in the first frame is at disposal.
Li \textit{et al}.~\cite{li2015robust} incorporated a convolutional
neural network (CNN) to visual tracking with multiple image cues as inputs.
In~\cite{zhou2014ensemble} an ensemble of deep networks have been combined by online boosting method for visual tracking.
However, due to the lack of sufficient training data, both methods have not demonstrated competitive results compared to state-of-the-art methods.
Another line of research resorts to numerous auxiliary data for offline training the deep networks, and then transfer the pre-trained model to online visual tracking.
Fan \textit{et al}.~\cite{fan2010human} proposed a human tracking
algorithm that learns a specific feature extractor with CNNs from an offline training set (about 20000
image pairs).
In~\cite{wang2013learning} Wang and Yeung proposed a deep learning
tracking method that uses stacked denoising autoencoder to learn the
generic features from a large number of auxiliary images (1 million
images).
Recently, Wang \textit{et al}.~\cite{wang2015video} employed a two-layer CNN to learn hierarchical features from auxiliary video sequences, which takes into account complicated motion transformations and appearance variations in visual tracking.
All these methods pay particular attention to offline learning an effective
feature extractor with a large amount of auxiliary data, yet do not
fully take into account the similar local structural and inner
geometric layout information among the targets over consequent frames,
which is handy and effective to discriminate the target from background for
visual tracking.
For instance, when tracking a face, the appearance and background in
consecutive frames change gradually, thereby providing strong similar
local structure and geometric layout in each tracked face (rather any
arbitrary pattern from a large dataset that covers numerous types of
objects).
%

%
\begin{figure}[t]
\begin{center}
 \includegraphics[width=0.8\linewidth]{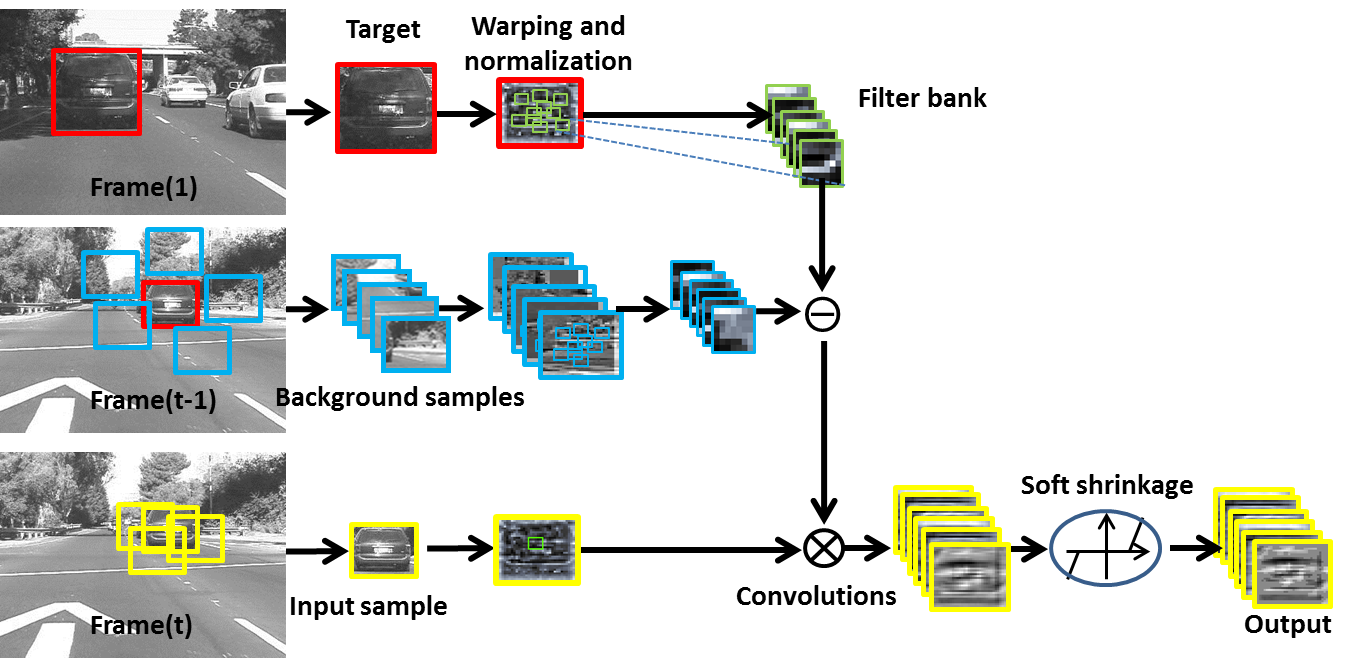}
\end{center}
   \caption{Overview of the proposed representation.
Input samples are warped into a canonical $32\times32$ images.
We first employ the $k$-means algorithm to extract a set of normalized local patches from the
warped target region in the first frame, and then extract a set of normalized local patches from the context region surrounding the target. We then use them as filters
to convolve each normalized sample extracted from subsequent frames,
resulting in a set of feature maps. Finally, the feature maps are de-noised by a soft shrinkage method, which results in a robust sparse representation.
}
\label{fig:representation}
\end{figure}
In this paper, we present a convolutional network based tracker (CNT),
which exploits the local structure and inner geometric layout
information of the target.
The proposed CNT has a surprisingly simple architecture, yet
effectively constructs a robust representation.
Figure~\ref{fig:representation} presents an overview of our method.
Different from the traditional CNNs~\cite{lecun1989backpropagation,lecun1998gradient} that combine three architectural ideas (i.e., local receptive fields, shared weights, and pooling with local average and subsampling) to address shift, scale, and distortion variance, our method does not include the pooling process due to the reasons as: first, high spatial resolution is needed to preserve the local structure of the target in visual tracking; second, the precise positions of the features lost by pooling play an important role to preserve the geometric layout of the target. The final image representation is global and sparse, which is a combination of some local feature maps. Such global image representation is built on the mid-level features~\cite{boureau2010learning}, which extract low-level information, but remain close to image-level information without any need of high-level structured image description. In~\cite{coates2011analysis} the simple $k$-means algorithm has also been employed to generate mid-level feature, which achieves promising performance.

The main contributions of this work are summarized as follows:
\begin{enumerate}
\item We present a convolutional network with a lightweight structure
  for visual tracking.
It is fully feed-forward and achieves fast speed
for online tracking  even on a CPU.
 \item Our method directly exploits the local structural and inner
   geometric layout information from data without manual tweaking,
which provides additional useful information besides appearance for visual tracking.
\item Our method achieves very competitive results  through evaluating on the CVPR2013 tracking
  benchmark dataset with 50 challenging
  videos~\cite{wu2013online} among 32 tracking algorithms including the state-of-the-art KCF~\cite{henriques2015high} and TGPR methods~\cite{gao2014transfer}.
In particular, it outperforms the recently proposed deep learning
tracker (DLT)~\cite{wang2013learning} (which requires offline training with
1 million auxiliary images) by a large margin (more than 10 percents in the AUC score of success rate), which shows the power of convolutional networks.
\end{enumerate}

\section{Related Work}
Our approach for object tracking is biologically inspired from recent
findings in neurophysiological studies.
First, we leverage convolution with predefined local filters (i.e., the
normalized image patches from the first frame) to extract the
high-order features, which is motivated by the HMAX model proposed by
Riesenhuber and Poggio~\cite{serre2007robust} that uses Garbor filters
instead.
Furthermore, we simply combine the local features without
changing their structures and spatial arrangements to generate a
global representation, which increases feature invariance while
maintaining specificity, thereby satisfying the two essential
requirements in cognitive task~\cite{riesenhuber1999hierarchical}.
In contrast, the HMAX model~\cite{serre2007robust} exploits a new
pooling mechanism with a maximum operation to enhance feature
invariance and specificity.
Second, our method owns a purely
feed-forward architecture, which is largely consistent with the
standard model of object recognition in primate
cortex~\cite{riesenhuber1999hierarchical} that focuses on the
capabilities of the ventral visual pathway in an immediate recognition
without the help of attention or other top-down effects.
The rapid performance of the human visual system suggests humans most
likely use feed-forward processing due to its simplicity.
Recently, psychophysical experiments
show that generic object tracking can be implemented in a low level
neural mechanism~\cite{mahadevan2012connections}, and hence our method
leverages a simple template matching scheme without using a high-level
object model.

 Most tracking methods emphasize on designing effective object
 representations~\cite{li2013survey}.
The holistic templates (i.e., raw
 image intensity) have been widely used in visual
 tracking~\cite{baker2004lucas,matthews2004template}.
Subsequently,
the online subspace-based method has been introduced to visual
tracking that handles appearance
variations well~\cite{ross2008incremental}.
Recently, Mei and
Ling~\cite{Mei_PAMI_2011} utilize a sparse representation of templates
to deal with the corrupted appearance of the target, which has been further improved
recently~\cite{bao2012real,zhang2012robust}.
Meanwhile, the local templates have attracted much attention in visual
tracking due to their robustness to partial occlusion and
deformation.
Adam \textit{et al}.~\cite{adam2006robust} use a set of local image patch
 histograms in a predefined grid structure to represent a
target object.
Kwon and Lee~\cite{kwon2009tracking} utilize a number of
local image patches to represent the target with an online scheme to
update their appearances and geometric relations.
Liu \textit{et al}.~\cite{liu2011robust} proposed a tracking method
that represents a target object by the histograms of sparse coding of
local patches.
However, in~\cite{liu2011robust} the local structural information of
the target has not been fully exploited. To address
this problem, Jia \textit{et al}.~\cite{jia2012visual} proposed an
alignment-pooling method to combine the histograms of
sparse coding.
Recently, the discriminative methods have been applied to
visual tracking due to the performance in which
a binary classifier is learned online to separate a target object from
the background.
Numerous learning methods have been developed to further improve
classifiers rather than image features based on
support vector machine (SVM) classifiers~\cite{Avidan_PAMI_2004},
structured output SVM~\cite{Hare_ICCV_2011}, online
boosting~\cite{Grabner_BMVC_2006},
P-N learning~\cite{Kalal_CVPR_2010}, multiple instance
learning~\cite{Babenko_PAMI_2011}, and some efficient hand-crafted
features are available off the shelf like the Haar-like
features~\cite{Grabner_BMVC_2006, Babenko_PAMI_2011,
  Hare_ICCV_2011,Zhang_ECCV_2012},
histograms~\cite{Grabner_BMVC_2006}, HOG descriptors~\cite{henriques2015high}, binary
features~\cite{Kalal_CVPR_2010}, and covariance descriptors~\cite{gao2014transfer}.

\begin{figure}[t]
\begin{center}
 \includegraphics[width=.8\linewidth]{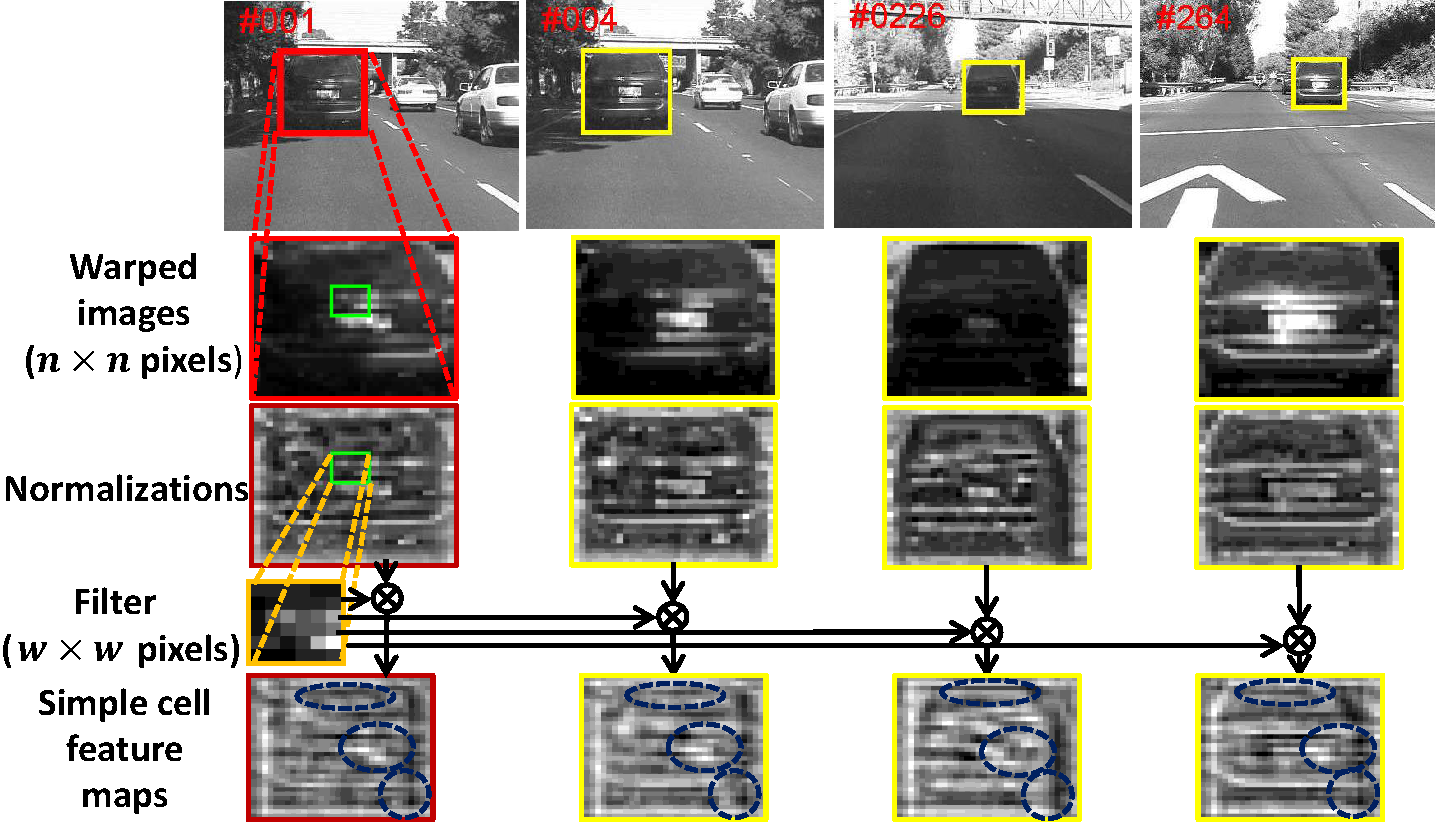}
\end{center}
   \caption{Although the target appearance changes significantly due to
     illumination changes and scale variations, the simple cell
     feature map can not only well preserve the local structure
(e.g., the regions in the dotted ellipses) of the target, but also
maintain its global geometric layout invariant to some degree.}
\label{fig:simplecellfeature}
\end{figure}
%
\section{Convolutional Networks for Tracking}
\subsection{Image Representation}
Given the target template, we develop a hierarchical representation
architecture with convolutional networks, including two separated
layers.
Figure~\ref{fig:representation} summarizes our
approach.
First, the local selective features are extracted with a
bank of filters convolving the input image at each position. Second,
the selective features are stacked together into a global
representation that is robust to appearance variations.
We refer these layers as the simple layer and the complex layer, respectively,
with analogy to the V1 simple and complex cells discovered by Hubel and
Wiesel~\cite{hubel1959receptive}.

%
\subsubsection{Preprocessing} We convert the input image to
grayscale and warp it to a fixed size with $n\times n$ pixels, denoted
as $\mathbf{I}\in \mathbb{R}^{n\times n}$.
We then densely sample a set of overlapping local image patches
$\mathcal{Y}=\{\mathbf{Y}_1,\ldots,\mathbf{Y}_l\}$ centered at each
pixel location inside the input image through sliding a window with
size $w\times w$ ($w$ is referred as the receptive field size),
where $\mathbf{Y}_i\in \mathbb{R}^{w\times w}$ is the $i$-th image patch and
$l=(n-w+1)\times (n-w+1)$. Each patch $\mathbf{Y}_i$ is preprocessed
by subtracting the mean and $\ell_2$ normalization that correspond to
local brightness and contrast normalization, respectively.

\subsubsection{Simple Layer} After preprocessing, we employ the $k$-means algorithm to
select a bank of patches
$\mathcal{F}^o=\{\mathbf{F}_1^o,\ldots,\mathbf{F}_d^o\}\subset \mathcal{Y}$
sampled from the object region in the first frame as fixed filters to extract our selective
simple cell features. Given the $i$-th filter
$\mathbf{F}_i^o\in\mathbb{R}^{w\times w}$, its response on the input
image $\mathbf{I}$ is denoted with a feature map
$\mathbf{S}_i^o\in\mathbb{R}^{(n-w+1)\times (n-w+1)} $, where
$\mathbf{S}_i^o=\mathbf{F}_i^o\otimes\mathbf{I}$.
As illustrated by
Figure~\ref{fig:simplecellfeature}, the filter $\mathbf{F}_i^o$ is
localized and selective that can extract the local structural features
(e.g., oriented edges, corners, endpoints), most of which are similar
despite the target appearance changing greatly.
Furthermore, the simple
cell feature maps have a similar geometric layout (see the bottom row
of Figure~\ref{fig:simplecellfeature}), which illustrates that the
local filter can extract useful information across the entire image,
and hence the global geometric layout information can also be
effectively exploited. Finally, the local filters can be referred
as a set of fixed local templates that encode stable information in
the first frame, thereby handling the drifting problem
effectively.
Similar strategy has been adopted
in~\cite{matthews2004template,liu2011robust,zhong2012robust},
where~\cite{matthews2004template} utilizes the template in the first
frame and the tracked result to update the template
while~\cite{liu2011robust,zhong2012robust} exploit a static dictionary
learned from the first frame to sparsely represent the tracked
target.

The background context surrounding the object provides useful information to discriminate the target from background.
As illustrated by Figure~\ref{fig:representation}, we choose $m$ background samples surrounding the object, and use the $k$-means algorithm to select
a bank of filters $\mathcal{F}_i^b=\{\mathbf{F}_{i,1}^b,\ldots,\mathbf{F}_{i,d}^b\}\subset \mathcal{Y}$ from the $i$-th background sample. Then, we employ the average pooling method to summarize each filter in $\mathcal{F}_i^b$ , which results in the background context filter set defined as $\mathcal{F}^b=\{\mathbf{F}_{1}^b=\frac{1}{m}\sum_{i=1}^m\mathbf{F}_{i,1}^b,\ldots,\mathbf{F}_{d}^b=\frac{1}{m}\sum_{i=1}^m\mathbf{F}_{i,d}^b\}$. Given the input image $\mathbf{I}$, the $i$-th background feature map is defined as $\mathbf{S}_i^b=\mathbf{F}_i^b\otimes\mathbf{I}$. Finally, the simple cell feature maps are defined as
\begin{equation}
\mathbf{S}_i=\mathbf{S}_i^o-\mathbf{S}_i^b=(\mathbf{F}_i^o-\mathbf{F}_i^b)\otimes\mathbf{I},i=1,\ldots,d.
\label{eq:simpleFeatureMap}
\end{equation}

\begin{figure}[t]
\begin{center}
 \includegraphics[width=1\linewidth]{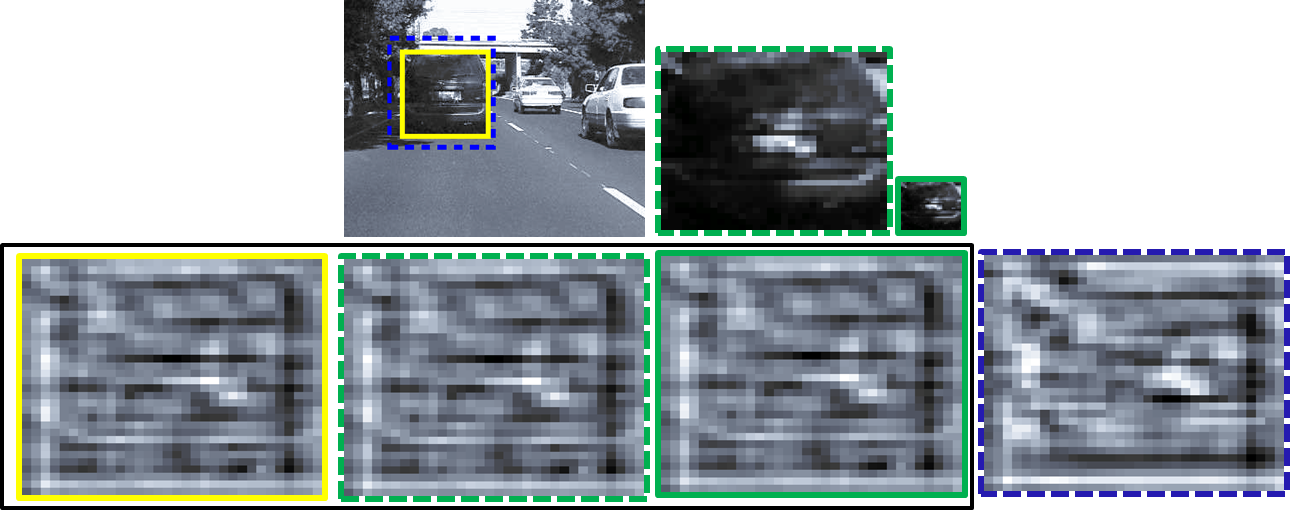}
\end{center}
   \caption{Illustration of the scale-invariant and shift-variant
     properties of the complex cell features. Scale invariance:
     although the scale of the target varies (top row),
     their corresponding simple feature maps (inside the black
     rectangle) have similar local structures and geometric layouts
     due to wrapping and normalization. Shift-variance:  the bottom
     right feature map generated by shifting the tracked target (in
     blue dotted rectangle) shows much difference
from the left ones due to inclusion of numerous background pixels.}
\label{fig:complexcellftr}
\end{figure}
%
\subsubsection{Complex Layer} The simple cell feature map
$\mathbf{S}_i$ simultaneously encodes the local structural and the
global geometric layout information of the target, thereby equipping
it with a good representation to handle appearance variations.
To further enhance the strength of this representation,
we construct a complex cell feature map
that is a 3D tensor $\mathbf{C}\in \mathbb{R}^{(n-w+1)\times
  (n-w+1)\times d}$, which simply stacks $d$ different simple cell
feature maps constructed with the filter set $\mathcal{F}$.
This layer is analogous to the pooling layers in the
CNNs~\cite{lecun1998gradient} and the HMAX
model~\cite{serre2007robust}: the CNNs utilize the local average and
subsampling operations while the HMAX model leverages the local maximum
scheme.

Both the CNNs and the HMAX model focus on learning shift-invariant
features that are useful for image classification and object
recognition~\cite{wang2013learning}, yet less effective for visual
tracking.
As illustrated in Figure~\ref{fig:complexcellftr}, if the complex
features are shift-invariant, both the blue dotted and the yellow
bounding boxes can be treated as the accurate tracking results,
leading to the location ambiguity problem. To overcome this problem,
in~\cite{Babenko_PAMI_2011} the multiple instance learning method is
introduced to visual tracking.
In contrast, the shift-variant
complex cell features make our method robust to location
ambiguity.
Furthermore, the complex cell features are more robust
to scale variation.
After warping the target at different scales to a
fixed size (e.g., $32\times 32$ pixels), the location of each useful
part in the target does not vary much in the warped images at this
abstracted view, and hence the complex cell features can preserve the
geometric layouts of the useful parts at different scales as well as
their local structures due to normalizing the wrapped target and the
local filters.

To make the map $\mathbf{C}$ robust to noise introduced by appearance variations, we utilize a sparse vector $\mathbf{c}$ to
approximate $\mathrm{vec}(\mathbf{C})$, which is obtained by minimizing the following objective function
\begin{equation}
\hat{\mathbf{c}}=\arg\min_{\mathbf{c}}\lambda\|\mathbf{c}\|_1^1+\frac{1}{2}\|\mathbf{c}-\mathrm{vec}(\mathbf{C})\|_2^2,
\label{eq:sparseFeatureMap}
\end{equation}
where $\mathrm{vec}(\mathbf{C})\in\mathbb{R}^{(n-w+1)^2d}$ is a column vector by concatenating all the elements in $\mathbf{C}$.
(\ref{eq:sparseFeatureMap}) has a closed form solution that can be readily achieved by a soft shrinkage function~\cite{elad2010role}
\begin{equation}
\hat{\mathbf{c}}=\mathrm{sign}(\mathrm{vec}(\mathbf{C}))\max(0,\mathrm{abs}(\mathrm{vec}(\mathbf{C}))-\lambda),
\end{equation}
where $\mathrm{sign}(\cdot)$ is a sign function, and we set $\lambda = \mathrm{median}(\mathrm{vec}(\mathbf{C}))$ that is the median value of $\mathrm{vec}(\mathbf{C})$ in our experiments, which can well adapt to target appearance variations during tracking.

\subsubsection{Model Update}
The sparse representation $\mathbf{c}$ in (\ref{eq:sparseFeatureMap}) serves as the target
template, which should be updated incrementally to accommodate
appearance changes over time for robust visual tracking.
We use a simple temporal low-pass filtering
method~\cite{zhang2014fast},
\begin{equation}
\mathbf{c}_t=(1-\rho)\mathbf{c}_{t-1}+\rho \mathbf{\hat{c}}_{t-1},
\label{eq:update}
\end{equation}
where $\rho$ is a learning parameter, $\mathbf{c}_t$ is the target
template at frame $t$ and $\mathbf{\hat{c}}_{t-1}$ is the
sparse representation of the tracked target at frame $t-1$. This simple
online update scheme not only accounts for rapid appearance
variations, but also alleviates drift problem due to retaining the
local filters in the first frame.

%
\subsubsection{Efficient Computation} The cost to compute the
target or background template $\mathbf{c}$ mainly includes preprocessing the local
patches in $\mathcal{Y}$ as well as convolving the input image
$\mathbf{I}$ with $d$ local filters in $\mathcal{F}$. However, the
operations of local normalization and mean subtraction when
preprocessing all patches can be reformulated as convolutions on the
input image~\cite{ben1999fast}.
Therefore, only the
convolution operations are needed when constructing the target
template, which can be efficiently computed by the fast Fourier
transforms (FFTs).
Furthermore, since the local filters are independent during tracking,
the convolutions can be easily parallelized, thereby
largely reducing the computational cost.

\subsection{Proposed Tracking Algorithm}
Our tracking algorithm is formulated within a particle filtering
framework.
Given the observation set
$\mathcal{O}_t=\{\mathbf{o}_1,\ldots,\mathbf{o}_t\}$ up to frame $t$,
our goal is to determine a posteriori probability
$p(\mathbf{s}_t|\mathcal{O}_t)$, which can be inferred by the Bayes'
theorem recursively
\begin{equation}
p(\mathbf{s}_t|\mathcal{O}_{t})\propto p(\mathbf{o}_t|\mathbf{s}_t)\int p(\mathbf{s}_t|\mathbf{s}_{t-1})p(\mathbf{s}_{t-1}|\mathcal{O}_{t-1})d\mathbf{s}_{t-1},
\label{eq:particle}
\end{equation}
where $\mathbf{s}_t=[x_t,y_t,s_t]^\top$ is the target state with translations $x_t, y_t$ and scale $s_t$,
$p(\mathbf{s}_t|\mathbf{s}_{t-1})$ is the motion model that predicts
the state $\mathbf{s}_t$ based on the previous state
$\mathbf{s}_{t-1}$, and $p(\mathbf{o}_t|\mathbf{s}_t)$ is the
observation model that estimates the likelihood of observation
$\mathbf{o}_t$ at the state $\mathbf{s}_t$ belonging to the target
category. We assume that the target state parameters are independent,
which are modeled by three scalar Gaussian distributions, and hence
the motion model can be formulated as a Brownian
motion~\cite{ross2008incremental}, i.e.,
$p(\mathbf{s}_t|\mathbf{s}_{t-1})= \mathcal{N}(\mathbf{s}_t
|\mathbf{s}_{t-1},\mathbf{\Sigma})$,
where $\mathbf{\Sigma}=\mbox{diag}(\sigma_x,\sigma_y,\sigma_s)$ is a diagonal
covariance matrix whose elements are the standard deviations of the
target state parameters. In visual tracking, the posterior probability
$p(\mathbf{s}_t|\mathcal{O}_t)$ in (\ref{eq:particle}) is approximated
by a particle filter in which $N$ particles
$\{\mathbf{s}_t^i\}_{i=1}^N$ are sampled with corresponding importance
weights $\{\pi_t^i\}_{i=1}^N$, where $\pi_t^i\varpropto
p(\mathbf{o}_t|\mathbf{s}_t^i)$. The optimal state is achieved by
maximizing the posteriori estimation over a set of $N$ particles
\begin{equation}
\mathbf{\hat{s}}_t=\arg\max_{\{\mathbf{s}_t^i\}_{i=1}^N}p(\mathbf{o}_t|\mathbf{s}_t^i)
p(\mathbf{s}_t^i|\mathbf{\hat{s}}_{t-1}).
\label{eq:map}
\end{equation}
The observation model $p(\mathbf{o}_t|\mathbf{s}_t^i)$
in~(\ref{eq:map}) plays a key role in robust tracking, and its
formulation in our method is
\begin{equation}
p(\mathbf{o}_t|\mathbf{s}^i_t)\varpropto e^{-\|\mathbf{c}_t
-\mathbf{c}_t^i\|_2^1},
\label{eq:observation}
\end{equation}
where $\mathbf{c}_t$ is the target template at frame $t$,
\begin{equation}
\mathbf{c}_t^i=\mathrm{vec}(\mathbf{C}_t^i)\odot\mathbf{w}
\end{equation}
 is the $i$-th candidate sample representation at
frame $t$ based on the complex cell features, where $\odot$ denotes the element-wise multiplication, and $\mathbf{w}$ is an indicator function whose element is defined as
\begin{equation}
 w_i=\left\{
\begin{aligned}
  &1, if~\mathbf{c}_t(i)\neq 0\\
  &0, else,
\end{aligned}
\right.
\end{equation}
where $\mathbf{c}_t(i)$ denotes the $i$-th element of $\mathbf{c}_t$.
With the incremental update scheme~(\ref{eq:update}), the observation model is
able to adapt to the target appearance variations while alleviating
the drift problem.

\begin{figure*}[t]
\begin{center}
\begin{tabular}{cc}
\includegraphics[width=0.35\linewidth]{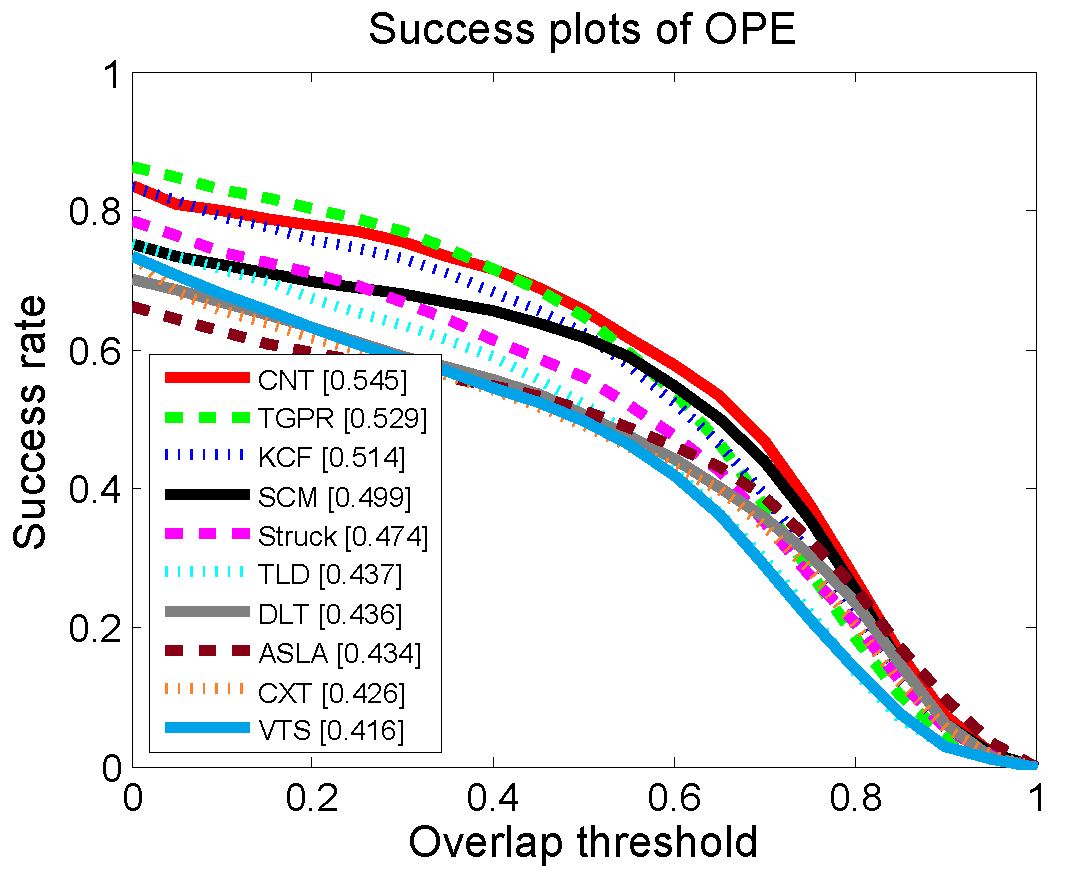}
\includegraphics[width=0.35\linewidth]{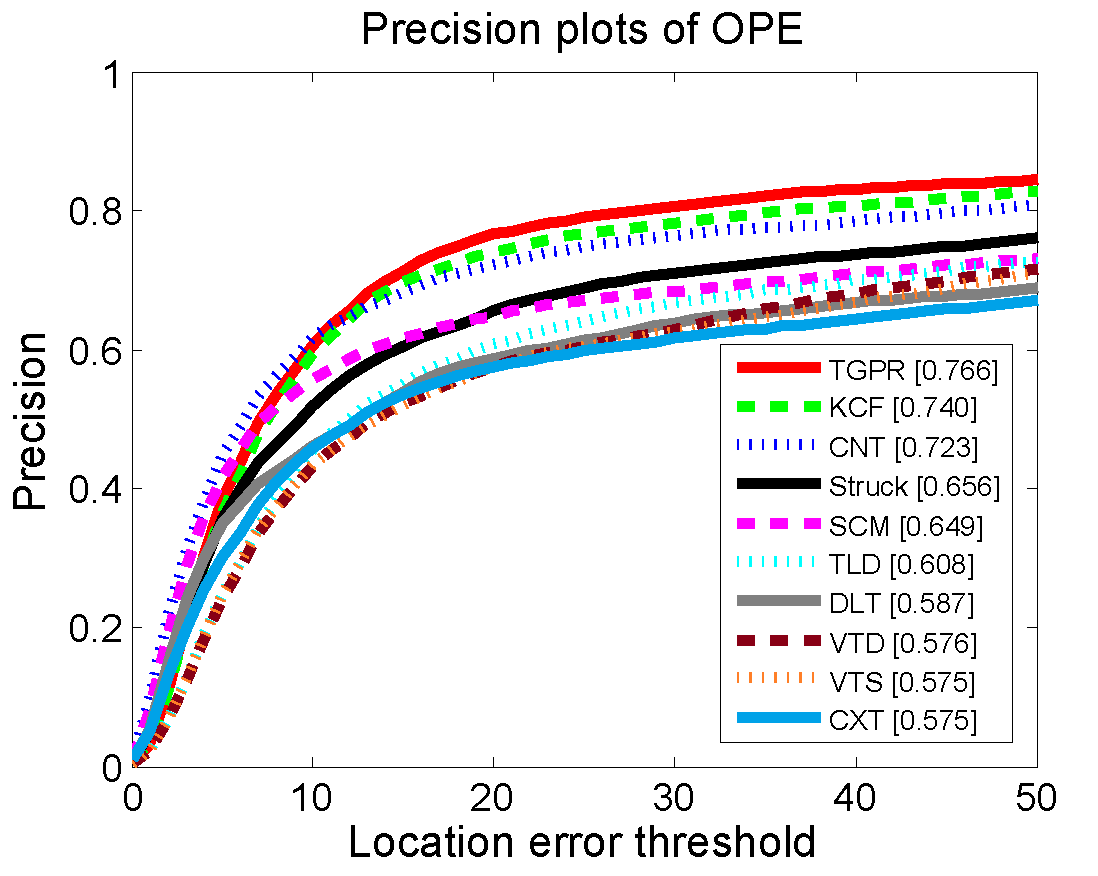}
\end{tabular}
\end{center}
   \caption{The success plots and precision plots of OPE for the top 10 trackers. The performance score of success plot is the AUC value while the performance score for each tracker is shown in the legend. The performance score of precession plot is at error threshold of 20 pixels while. Best viewed on color display.}
\label{fig:overallperformance}
\end{figure*}
\section{Experiments}
\subsection{Experimental Setup}
The proposed CNT is implemented in MATLAB and
runs at $5$ frames per second on a PC with Intel i7 3770
CPU (3.4 GHz).
Each color video is converted to grayscale, and the
state of the target (i.e., size and location) in the first frame is
given by the ground truth.
The size of the warped image $n\times n$ is
set to $n=32$. The receptive field size $w\times w$ is set to
$w=6$. The number of filters is set to $d=100$. The learning parameter
$\rho$ in~(\ref{eq:update}) is set to $0.95$ and the template is
updated every frame. The standard deviations of the target state of
the particle filter are set to $\sigma_x=4$, $\sigma_y=4$, and
$\sigma_s=0.01$, and $N=600$ particles are used. The parameters are
fixed for all experiments.
The source code will be made available to the public.

\subsection{Evaluation Metric}
 We use the CVPR2013 tracking benchmark
dataset and code library~\cite{wu2013online}, which includes 29
trackers and 50 fully-annotated videos (more than 29,000 frames).
Furthermore, we also add the results of two state-of-the-art trackers including the KCF~\cite{henriques2015high} and TGPR methods~\cite{gao2014transfer}, and one representative deep learning based tracker DLT~\cite{wang2013learning}.
To  better evaluate and analyze the strength and weakness of the
tracking approaches, the videos are categorized with 11 attributes
based on different challenging factors, including illumination
variation, scale variation, occlusion, deformation, motion blur, fast
motion, in-plane rotation, out-of-plane rotation, out-of-view,
background clutters, and low resolution.

For quantitative comparison, we
employ the success plot and the precision plot used
in~\cite{wu2013online}.
The success plot is based on the overlap ratio
that is  $S=\mathrm{Area}(B_T\cap B_G)/\mathrm{Area}(B_T\cup B_G)$,
where $B_T$ is the tracked bounding box and $B_G$ denotes the ground
truth. The success plot shows the percentage of frames with $S>t_0$
throughout all threshold $t_0\in [0,1]$. The area under curve (AUC) of
each success plot serves as the second measure to rank the tracking
algorithms.
Meanwhile, the precision plot illustrates the percentage
of frames whose tracked locations are within the given threshold
distance to the ground truth.
A representative precision score with
the threshold equal to $20$ pixels is used to rank the
trackers.
We report the results of one-pass evaluation (OPE)~\cite{wu2013online}
based on the average success and precision rate given the ground truth
target state in the first frame.
For presentation clarity, we only present the top
10 algorithms in each plot. The demonstrated evaluated trackers include the proposed CNT,
KCF~\cite{henriques2015high}, TGPR~\cite{gao2014transfer}, Struck~\cite{Hare_ICCV_2011}, SCM~\cite{zhong2012robust},
TLD~\cite{Kalal_CVPR_2010}, DLT~\cite{wang2013learning},
VTD~\cite{Kwon_CVPR_2010}, VTS~\cite{kwon2011tracking},
CXT~\cite{dinh2011context}, CSK~\cite{henriques2012circulant},
ASLA~\cite{jia2012visual}, DFT~\cite{sevilla2012distribution},
LSK~\cite{liu2011robust}, CPF~\cite{perez2002color},
LOT~\cite{oron2012locally}, TM-V~\cite{collins2005open},
KMS~\cite{comaniciu2003kernel}, L1APG~\cite{bao2012real},
MTT~\cite{zhang2012robust}, MIL~\cite{Babenko_PAMI_2011},
OAB~\cite{Grabner_BMVC_2006}, and SemiT~\cite{grabner2008semi}.

\subsection{Quantitative Comparisons}
\subsubsection{Overall Performance}
Figure~\ref{fig:overallperformance} illustrates the overall
performance of the top 10 performing tracking algorithms in terms of
success plot and precision plot.
Note that all the plots are
automatically generated by the code library supported by the
benchmark evaluation~\cite{wu2013online}, and the results of
KCF~\cite{henriques2015high}, TGPR~\cite{gao2014transfer}, and DLT~\cite{wang2013learning} are provided by the authors.
The proposed
CNT ranks 1st based on the success rate while 3rd based on the precision rate:
in the success plot, the
proposed CNT achieves the AUC of 0.545, which outperforms DLT by $10.9\%$.
Meanwhile, in the precision plot, its precision score is 0.723, closely following TGPR 0.766 and KCF 0.740, but
outperforms significantly DLT by $14.5\%$.
Note that
the proposed CNT exploits only the simple sparse image representation that encodes
the local structural and geometric layout information of the target, yet achieves
competitive performance to Struck and SCM that utilize useful
background information to train discriminative
classifiers.
Furthermore, even using only specific target
information from the first frame without further learning with auxiliary training data, CNT can
still outperform DLT by a wide margin (more than 10 percents in terms of both success and precision rates), showing that the generic
features offline learned from numerous auxiliary data may not adapt
well to target appearance variations in visual tracking.
\subsubsection{Attribute-based Performance}
To facilitate analyzing the strength and weakness of the proposed
algorithm, following~\cite{wu2013online}, we further evaluate the
trackers on videos with 11
attributes.
\begin{figure*}[t]
\begin{center}
\begin{tabular}{cccc}
\includegraphics[width=0.25\linewidth]{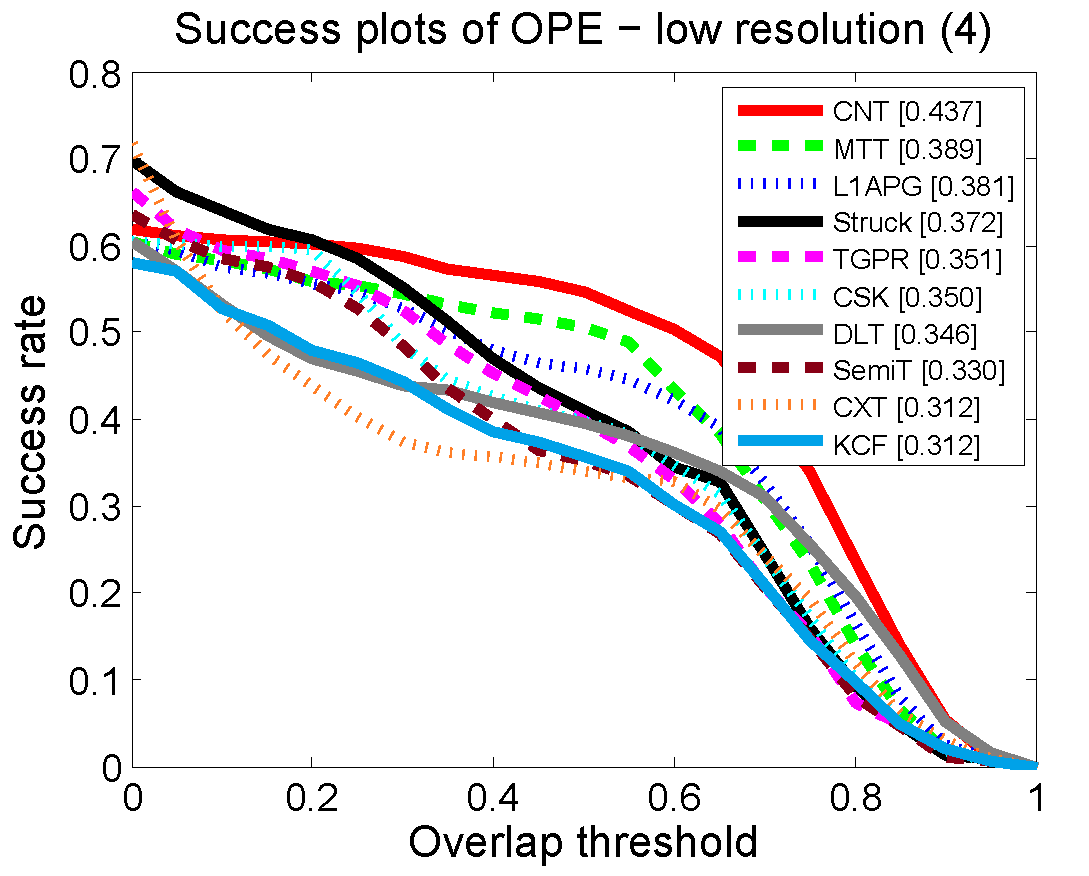}
\includegraphics[width=0.25\linewidth]{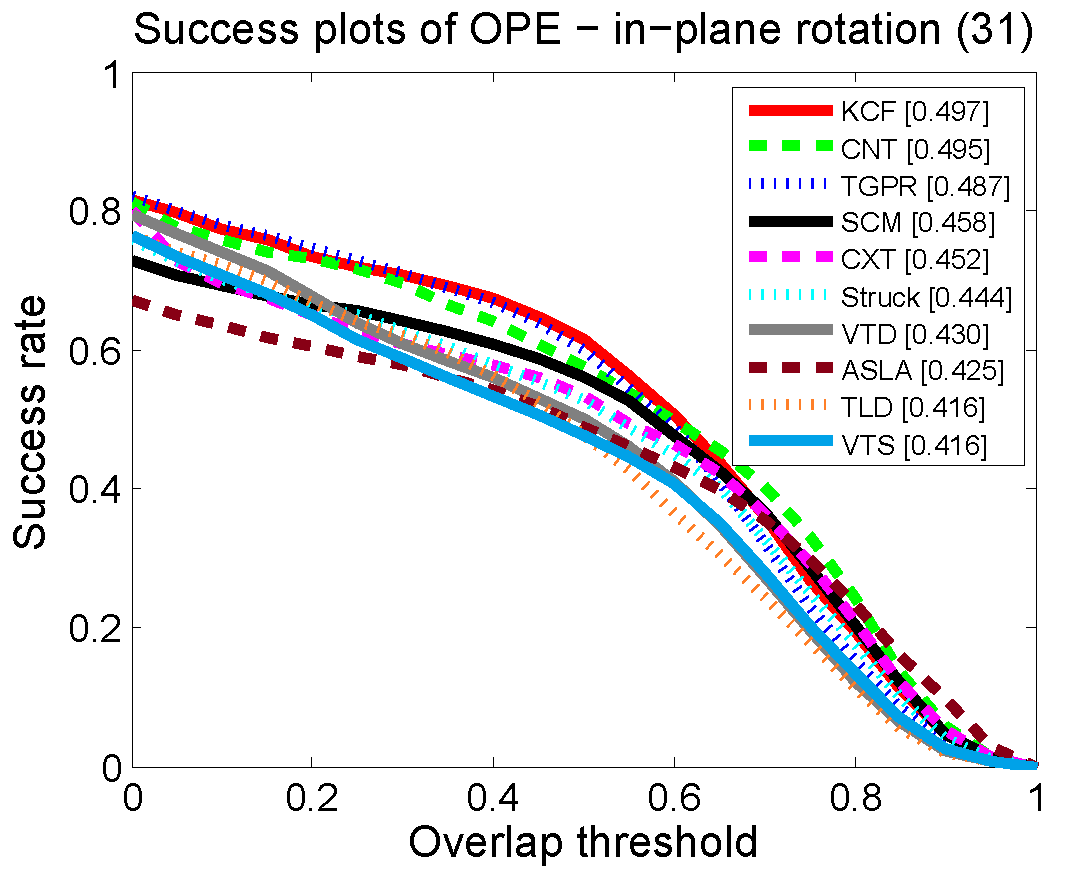}
 \includegraphics[width=0.25\linewidth]{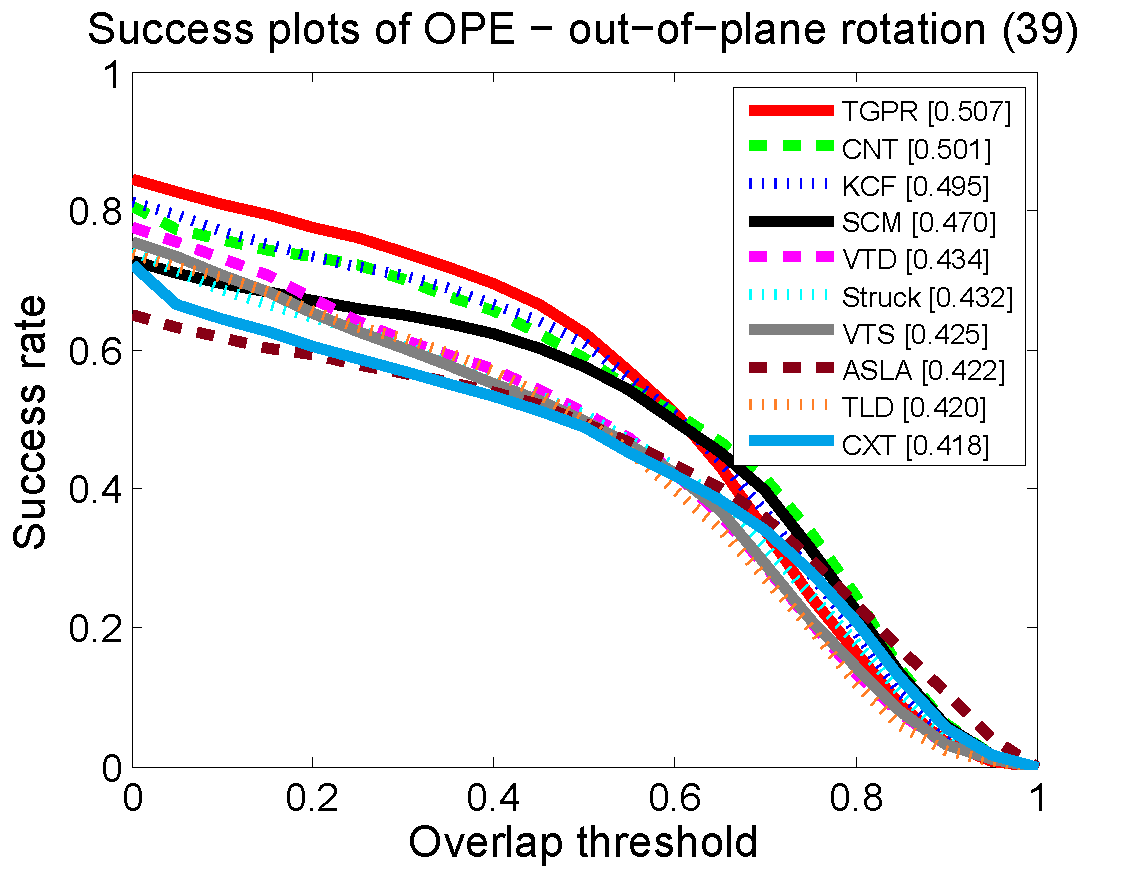}
 \includegraphics[width=0.25\linewidth]{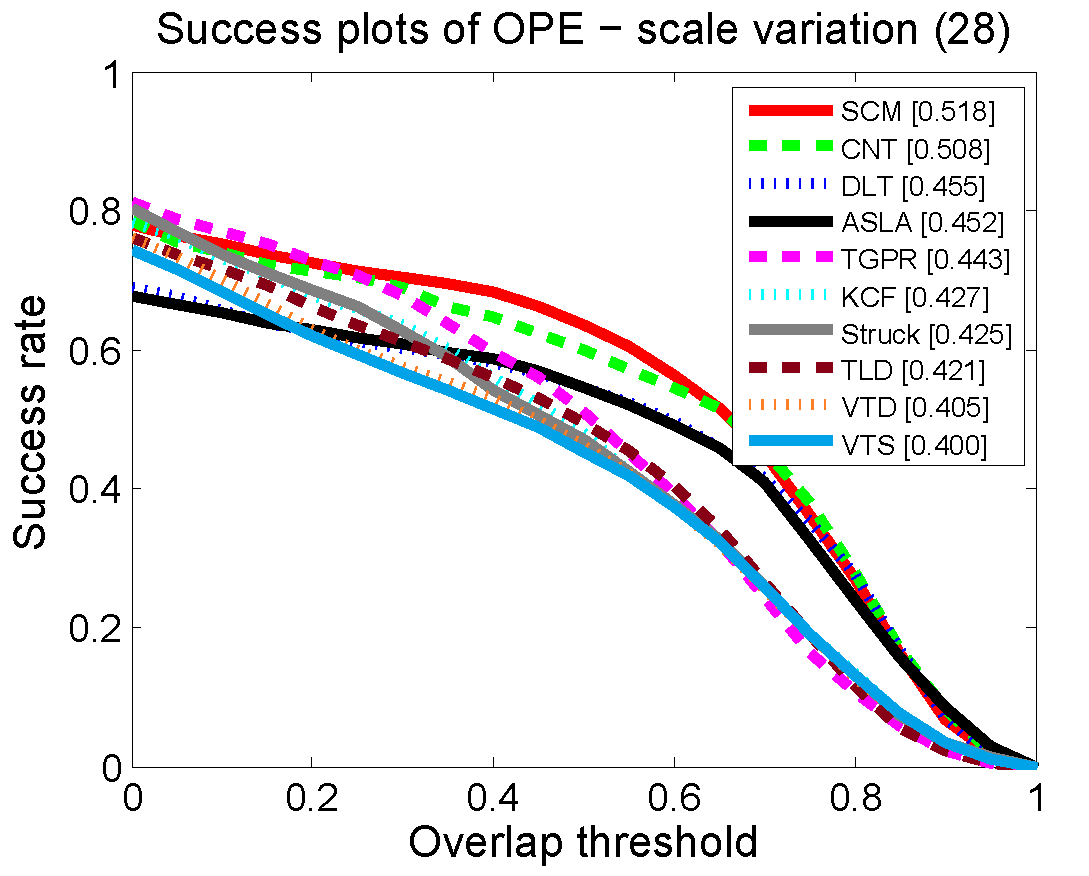}\\
 \includegraphics[width=0.25\linewidth]{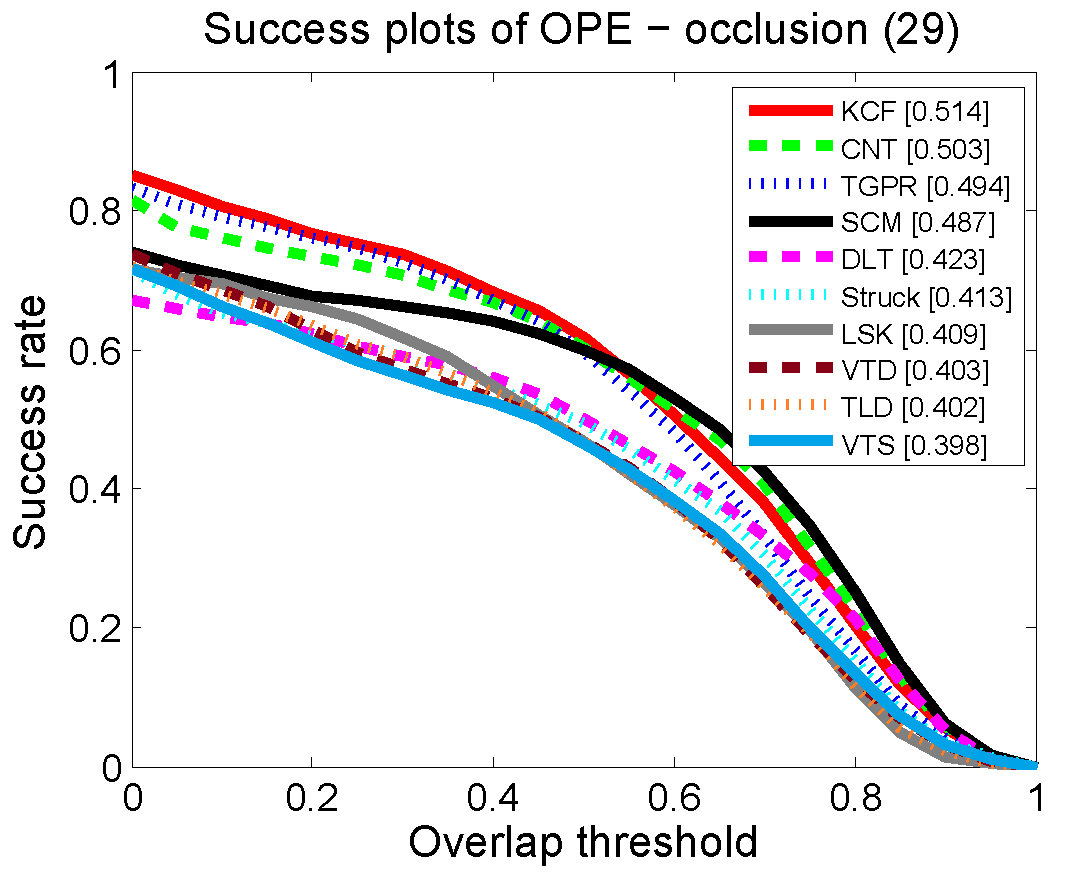}
 \includegraphics[width=0.25\linewidth]{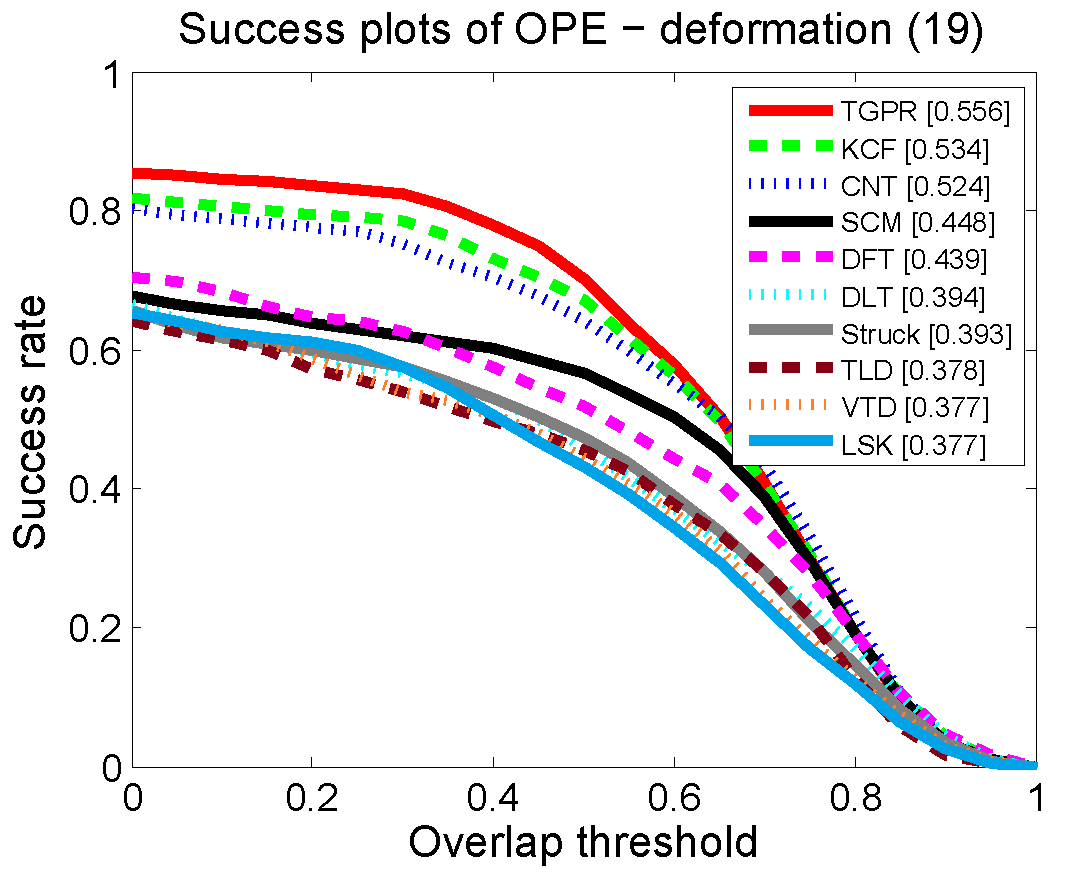}
 \includegraphics[width=0.25\linewidth]{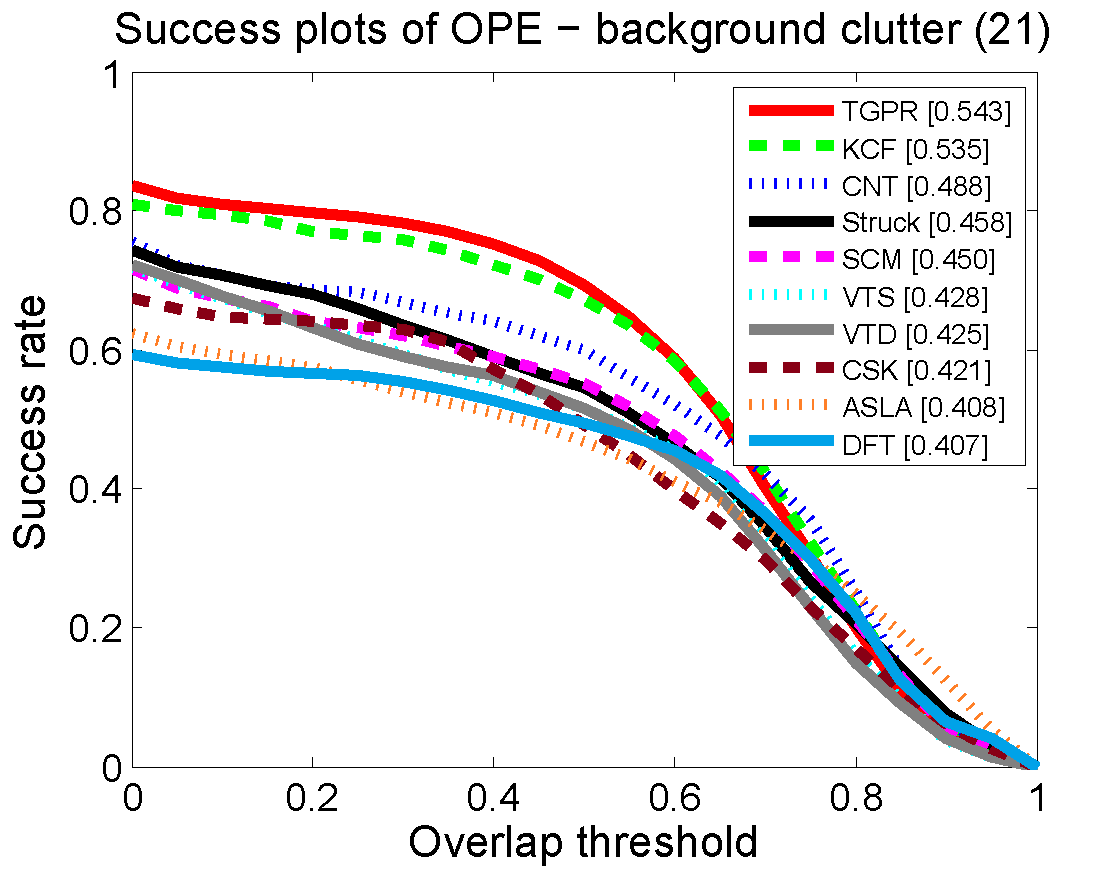}
 \includegraphics[width=0.25\linewidth]{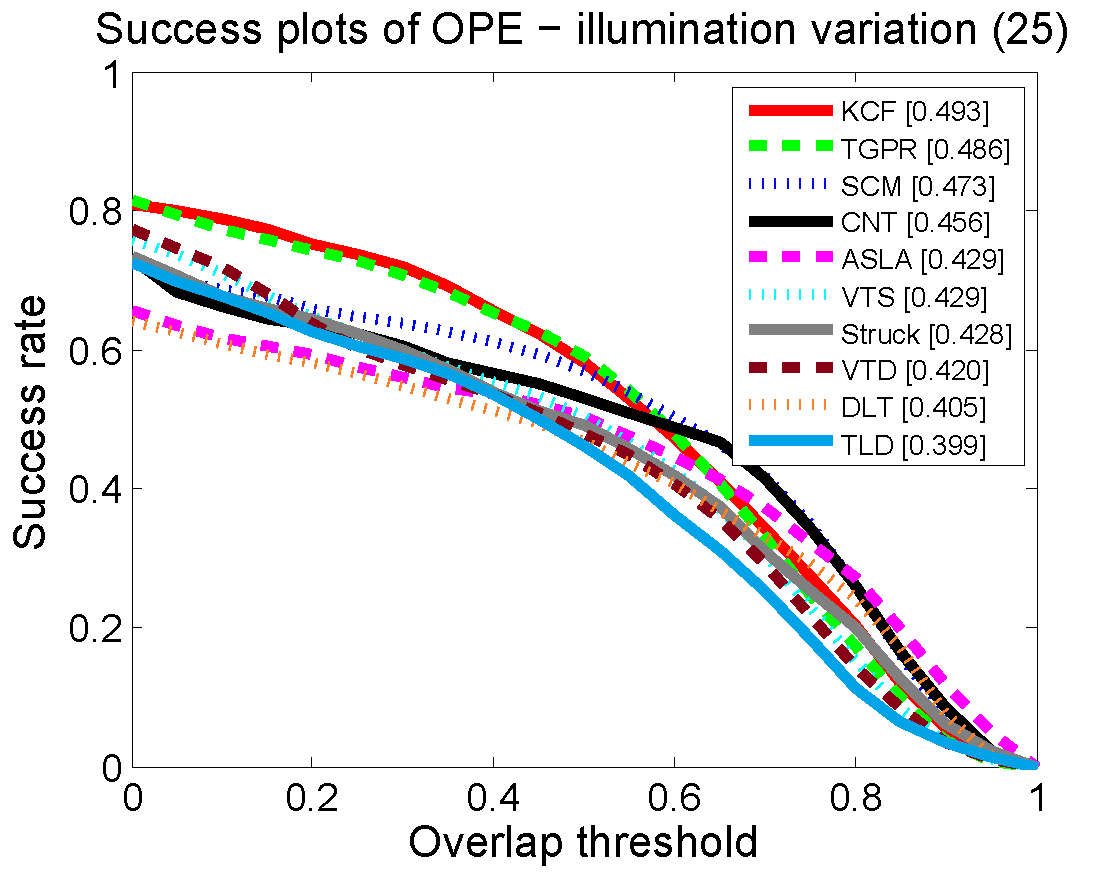}\\
 \includegraphics[width=0.25\linewidth]{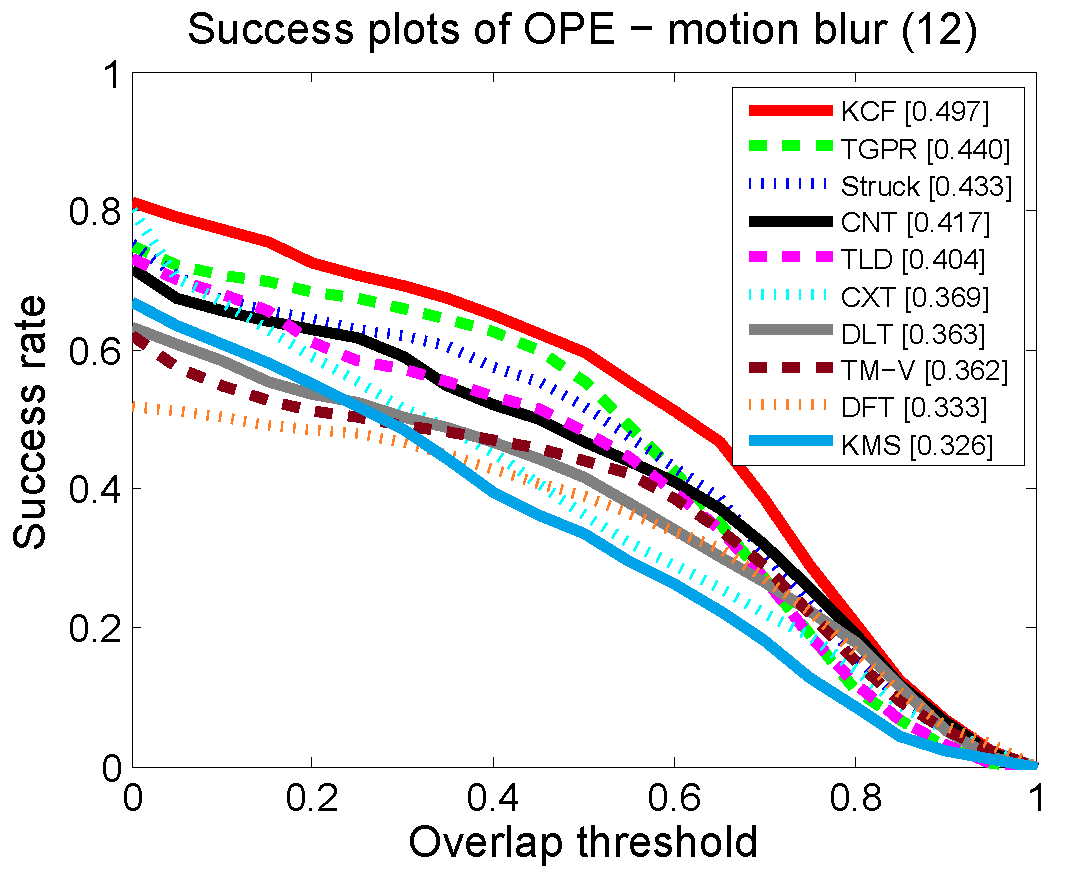}
 \includegraphics[width=0.25\linewidth]{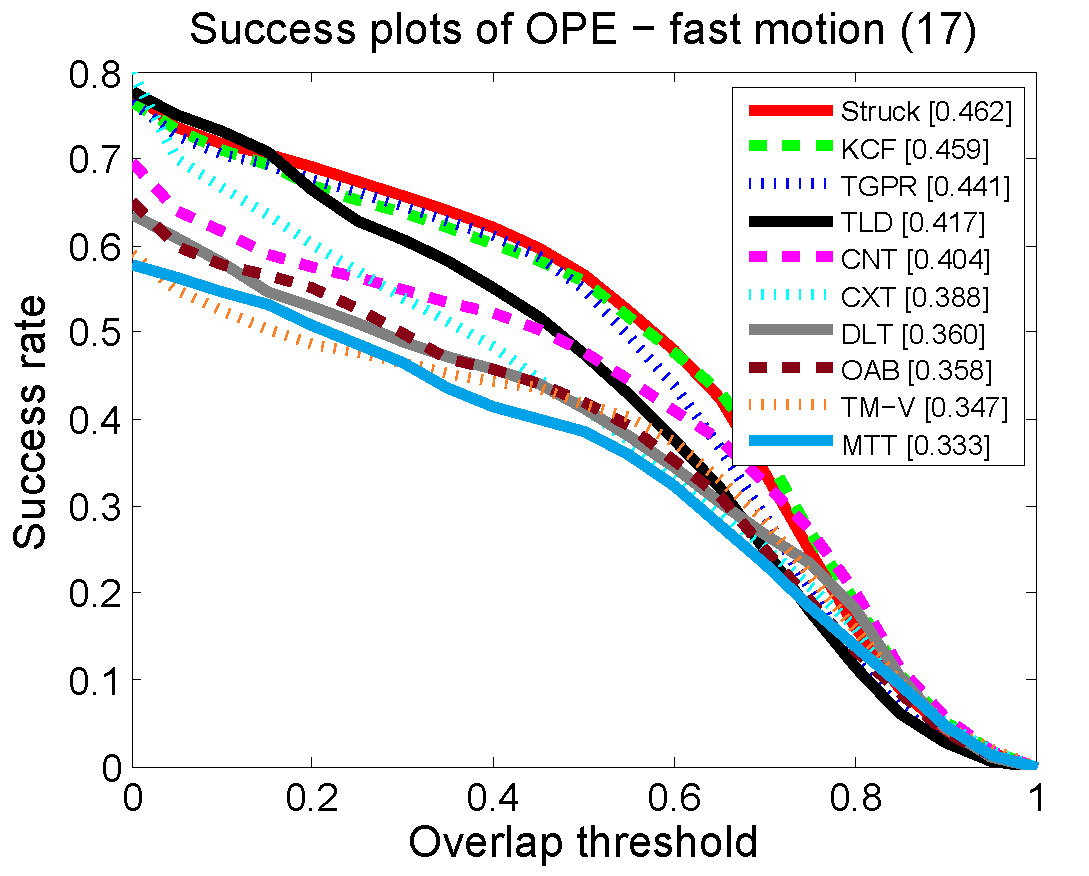}
 \includegraphics[width=0.25\linewidth]{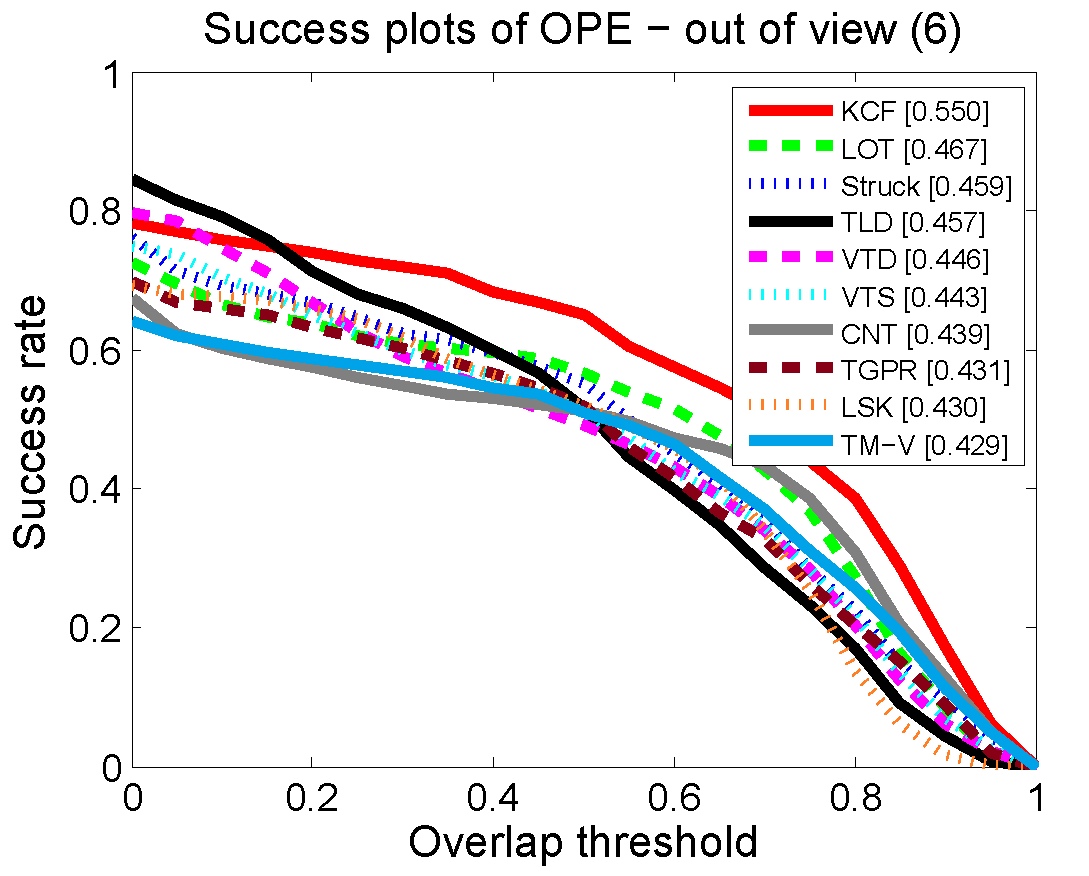}
\end{tabular}
\end{center}
   \caption{The success plots of videos with different attributes. Best viewed on color display.}
\label{fig:attributeSuccess}
\end{figure*}
\begin{figure*}[t]
\begin{center}
\begin{tabular}{cccc}
\includegraphics[width=0.25\linewidth]{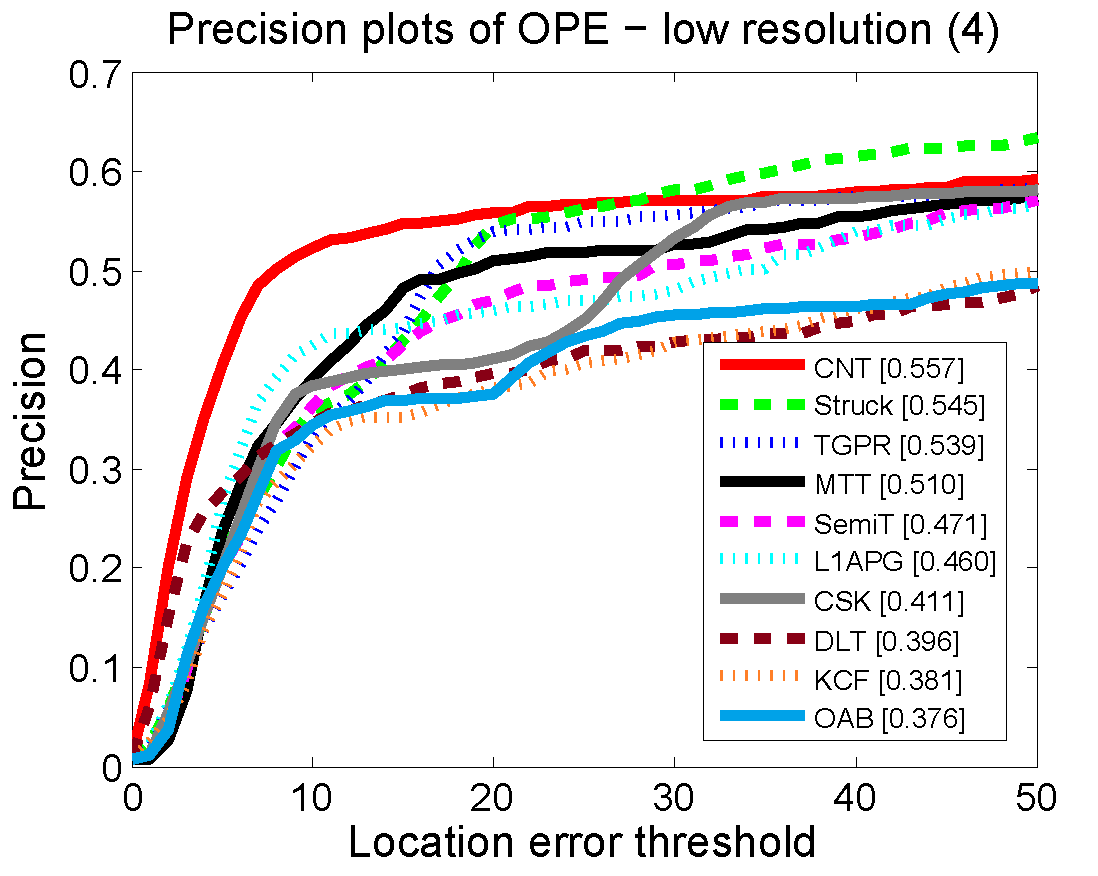}
\includegraphics[width=0.25\linewidth]{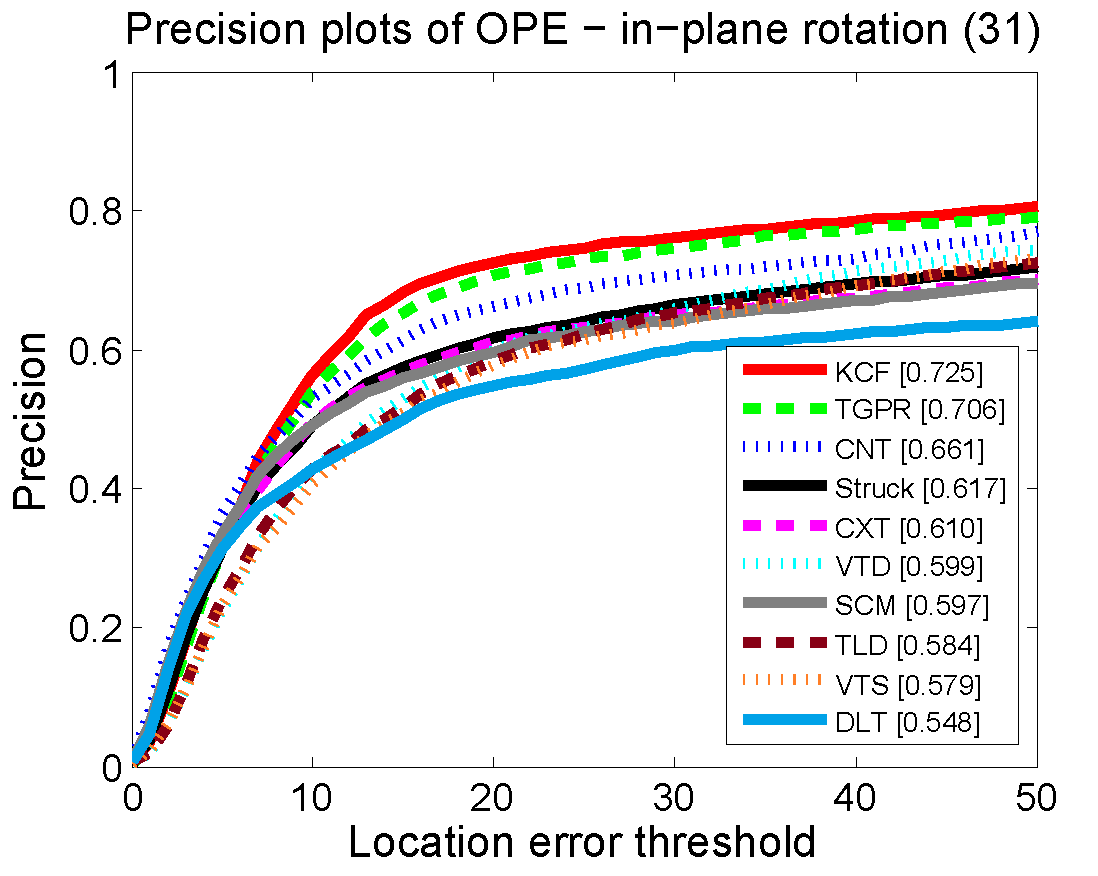}
 \includegraphics[width=0.25\linewidth]{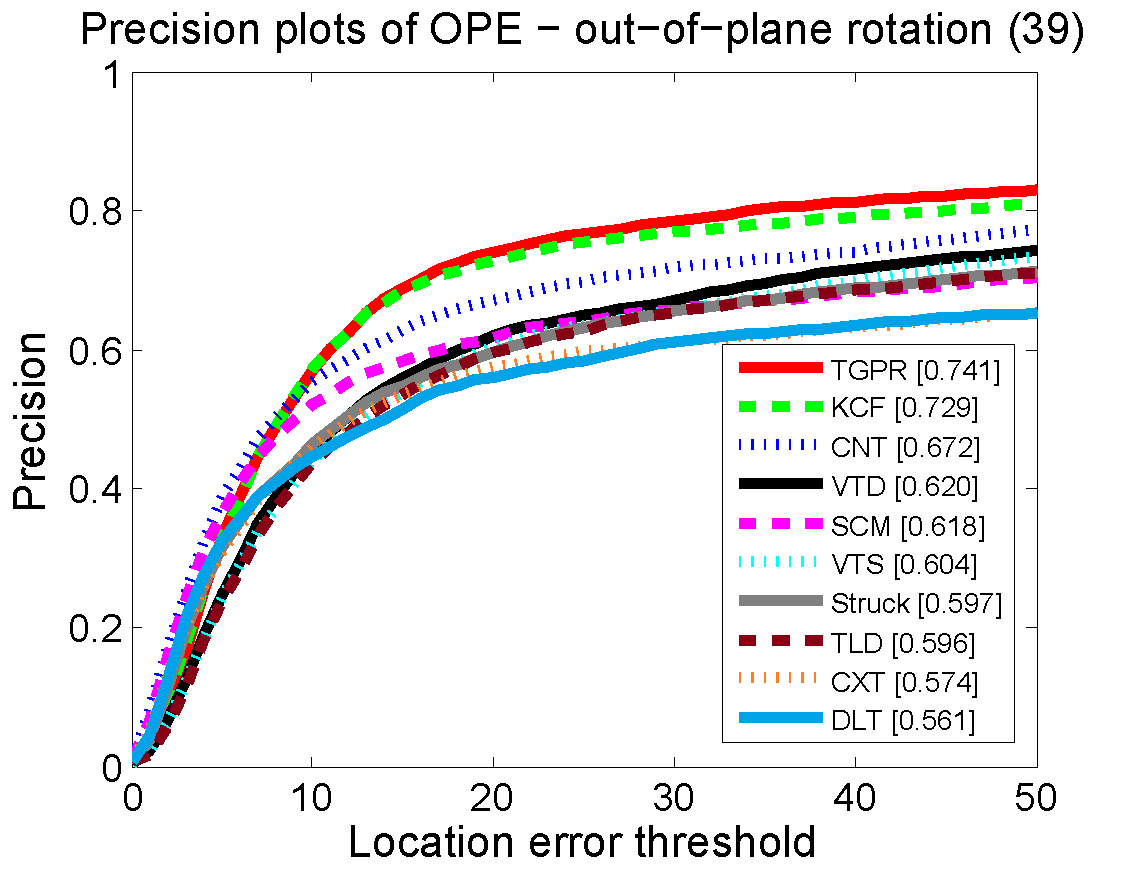}
 \includegraphics[width=0.25\linewidth]{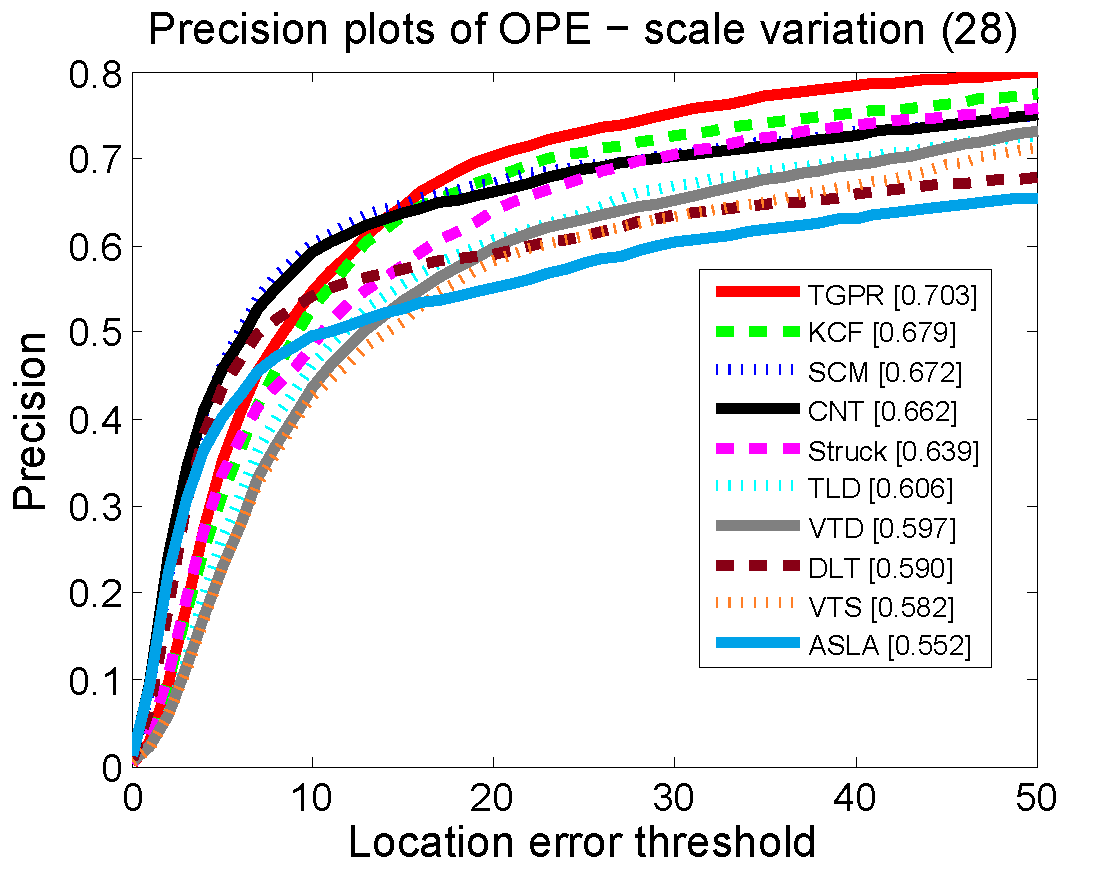}\\
 \includegraphics[width=0.25\linewidth]{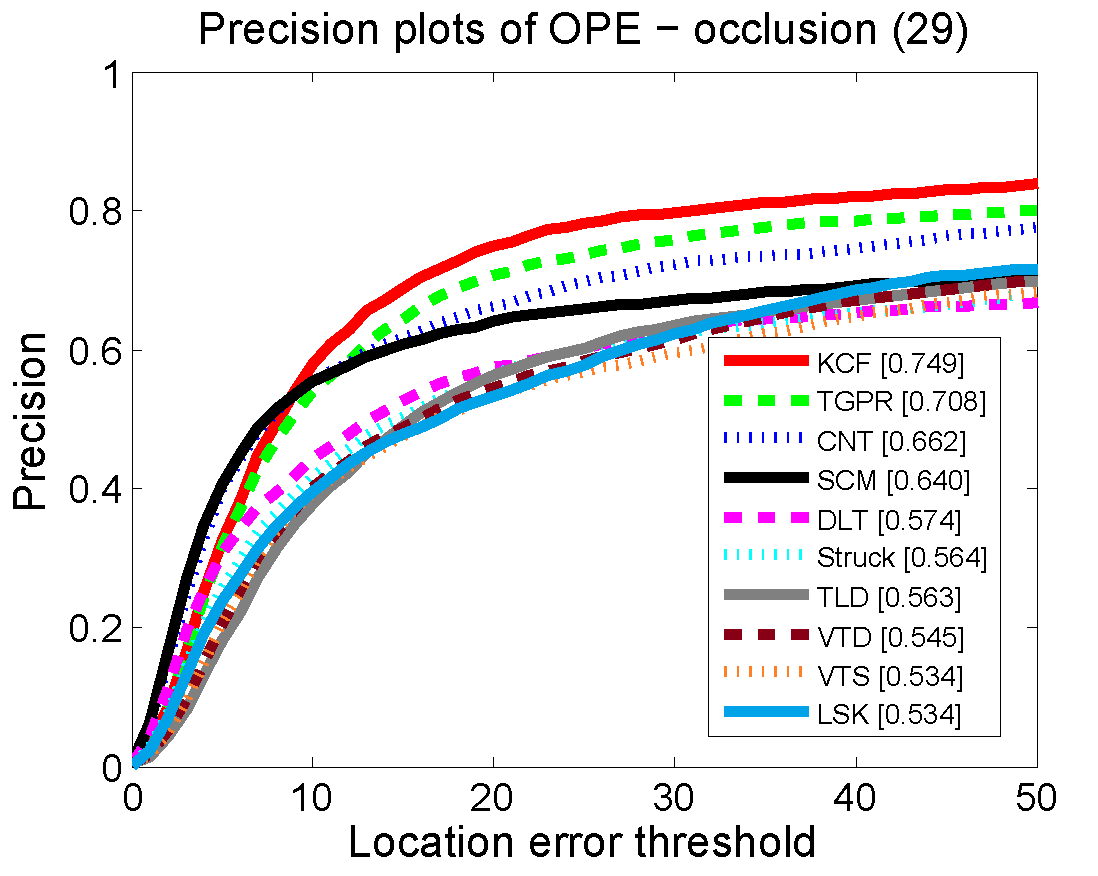}
 \includegraphics[width=0.25\linewidth]{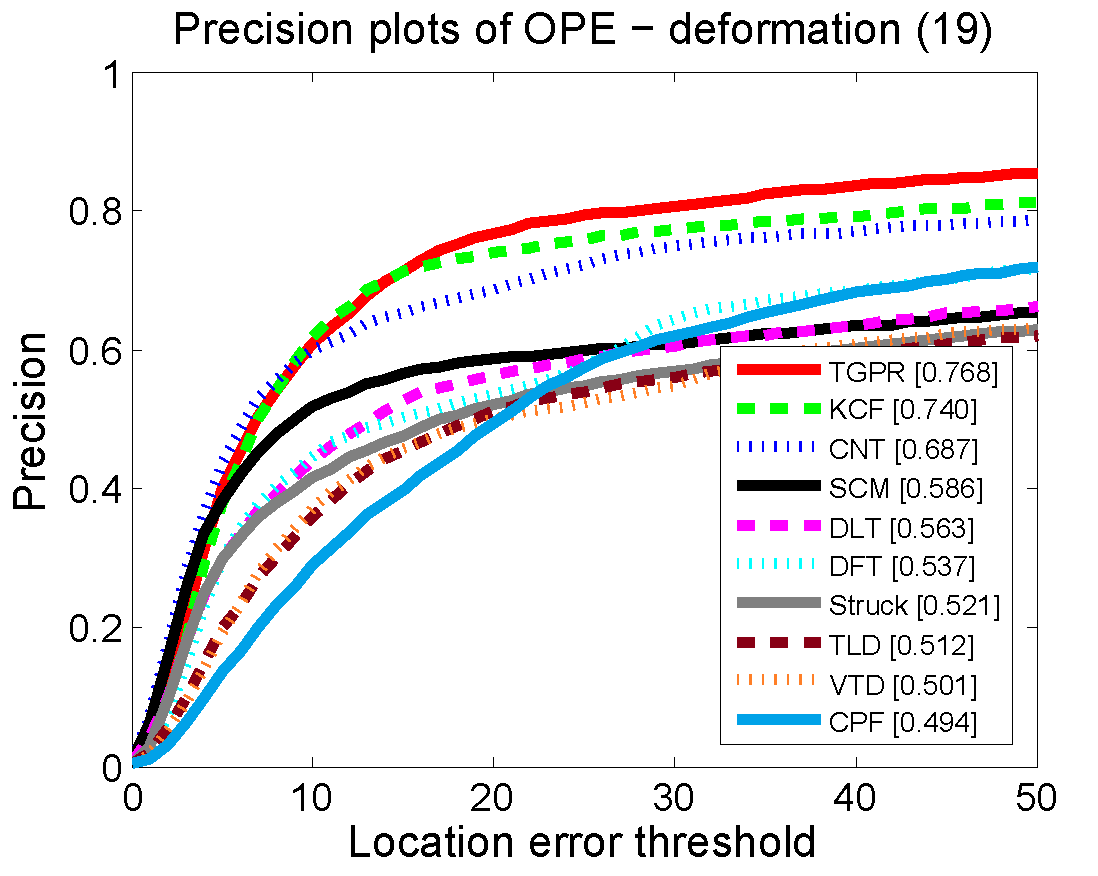}
 \includegraphics[width=0.25\linewidth]{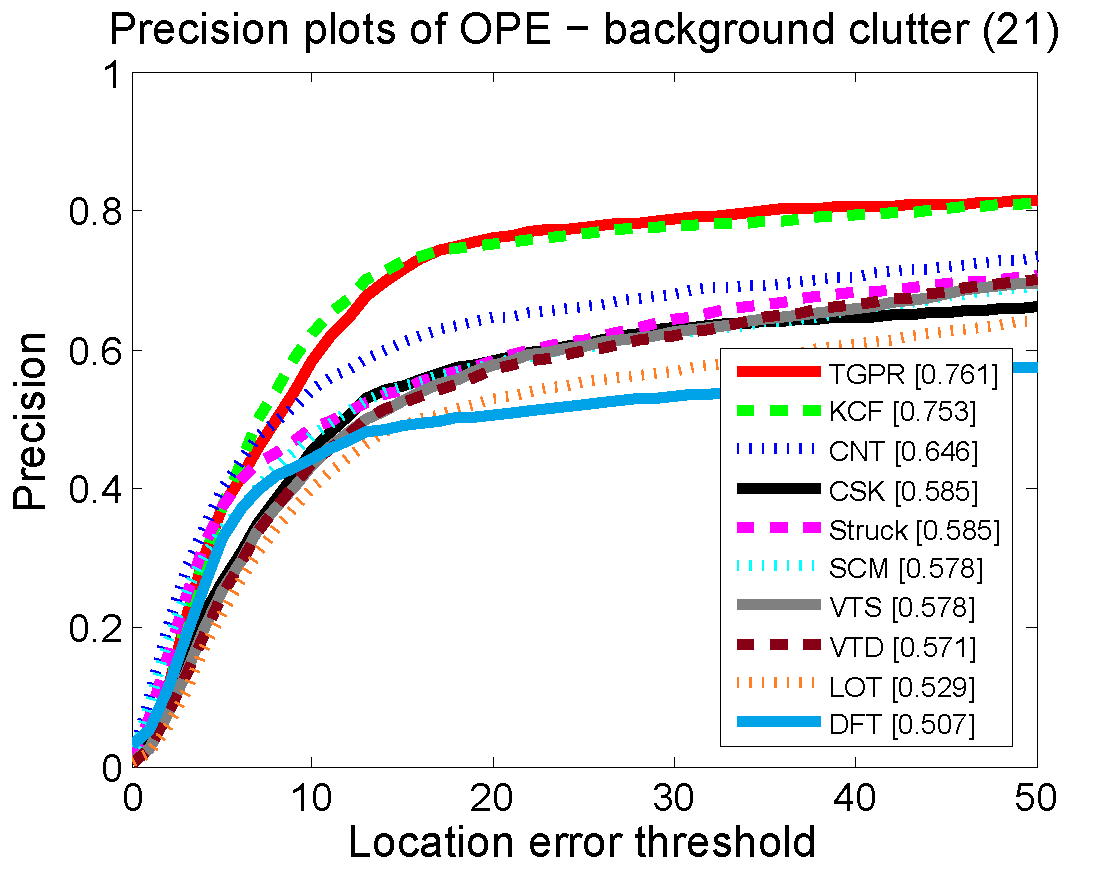}
 \includegraphics[width=0.25\linewidth]{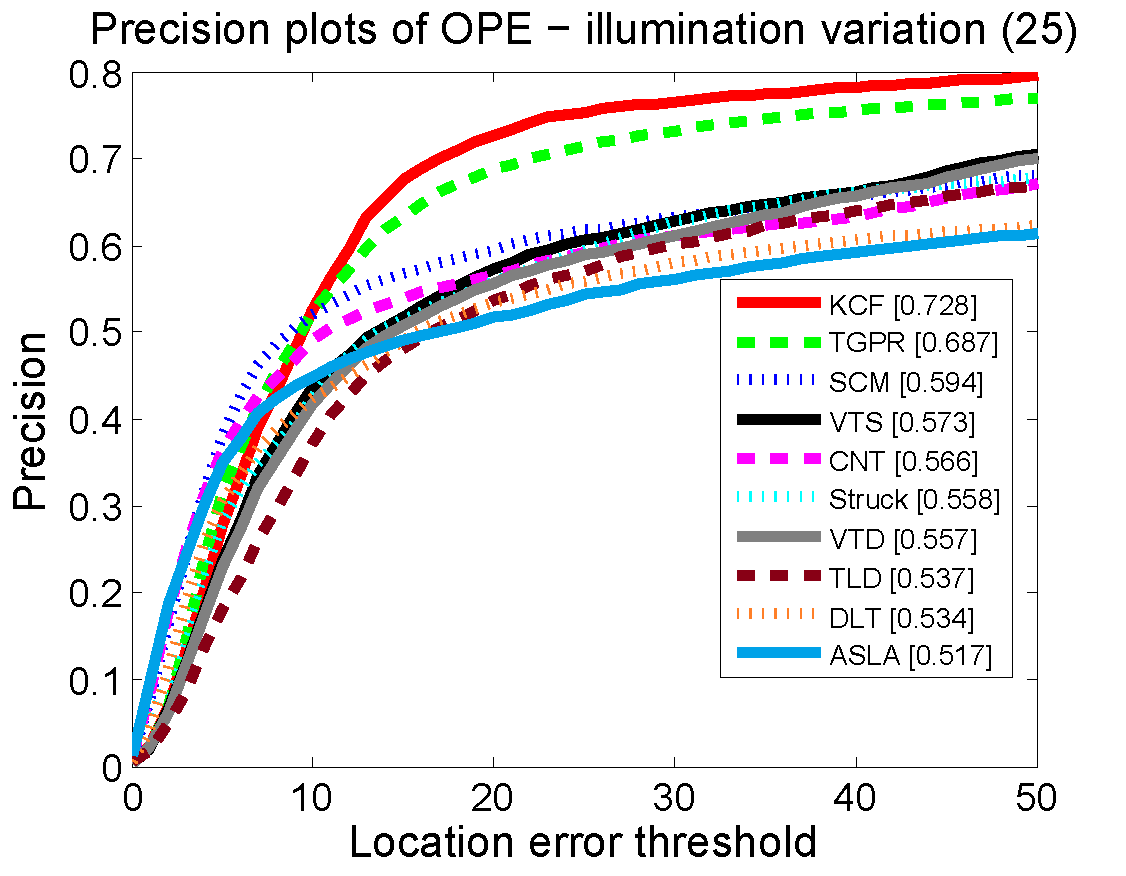}\\
  \includegraphics[width=0.25\linewidth]{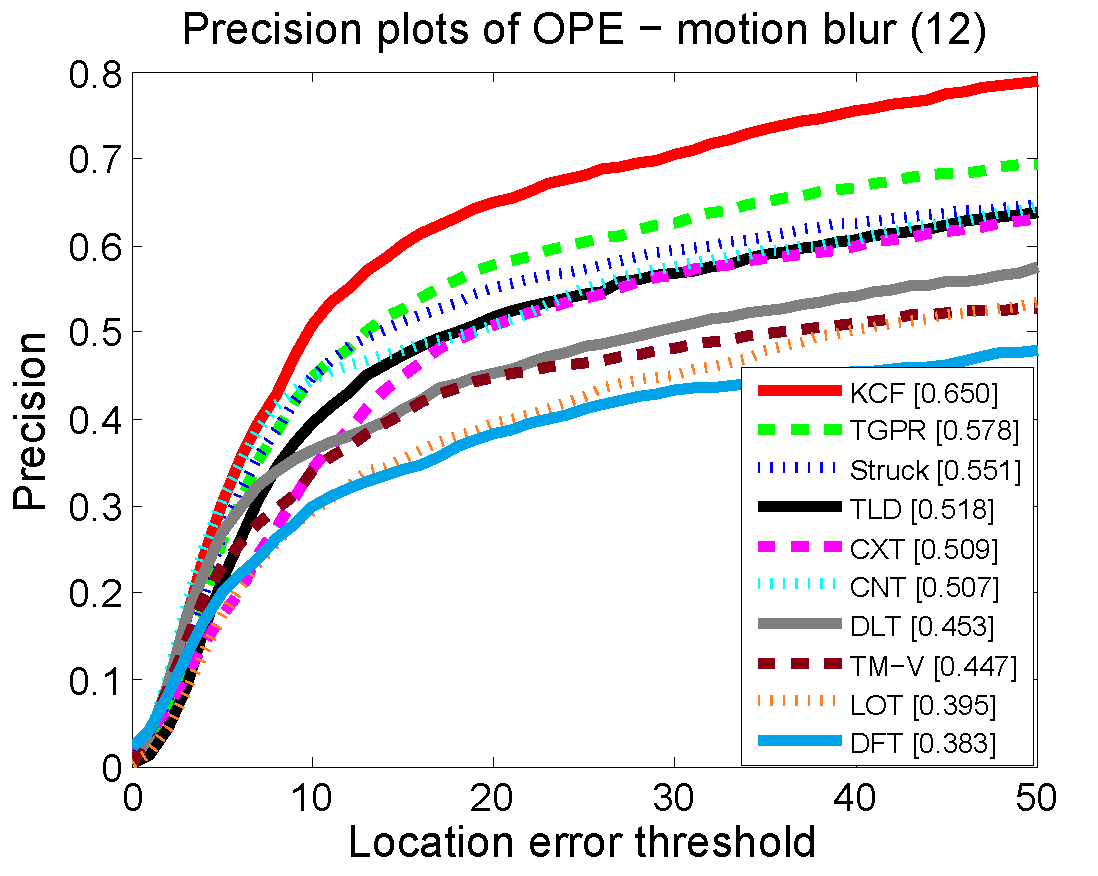}
 \includegraphics[width=0.25\linewidth]{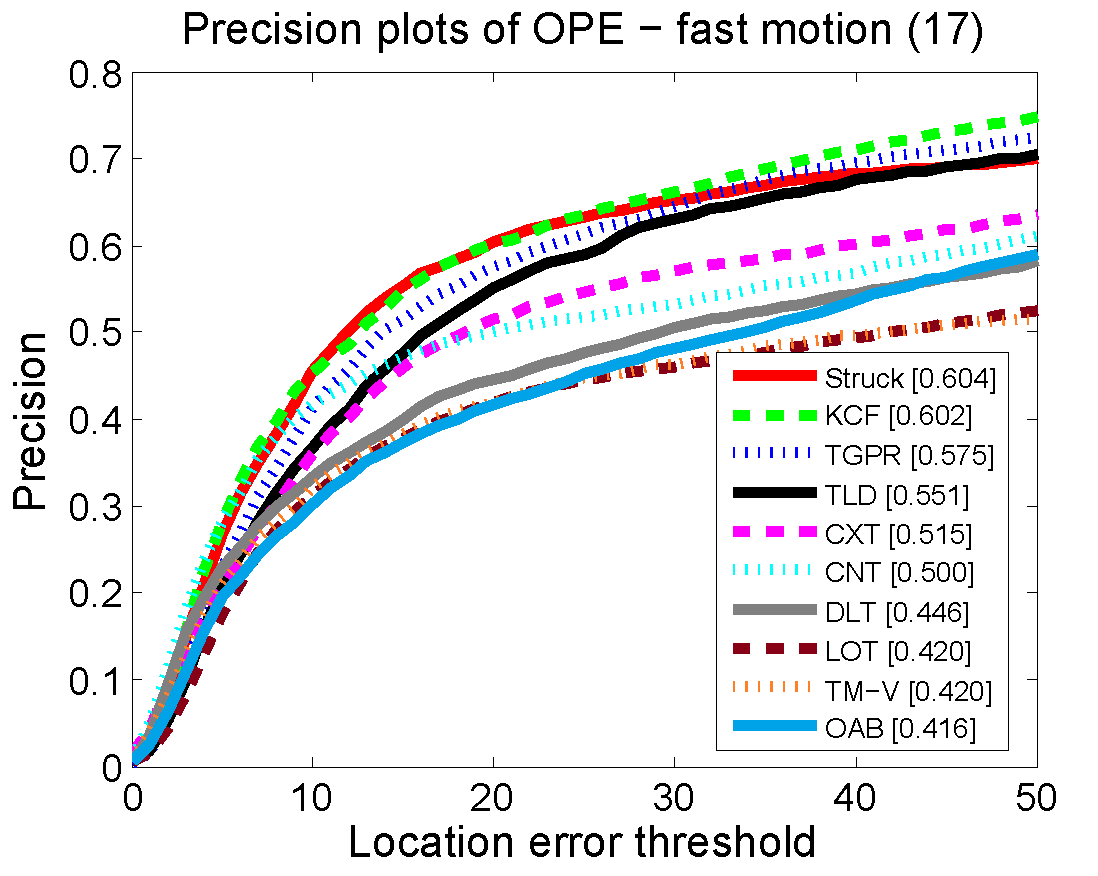}
 \includegraphics[width=0.25\linewidth]{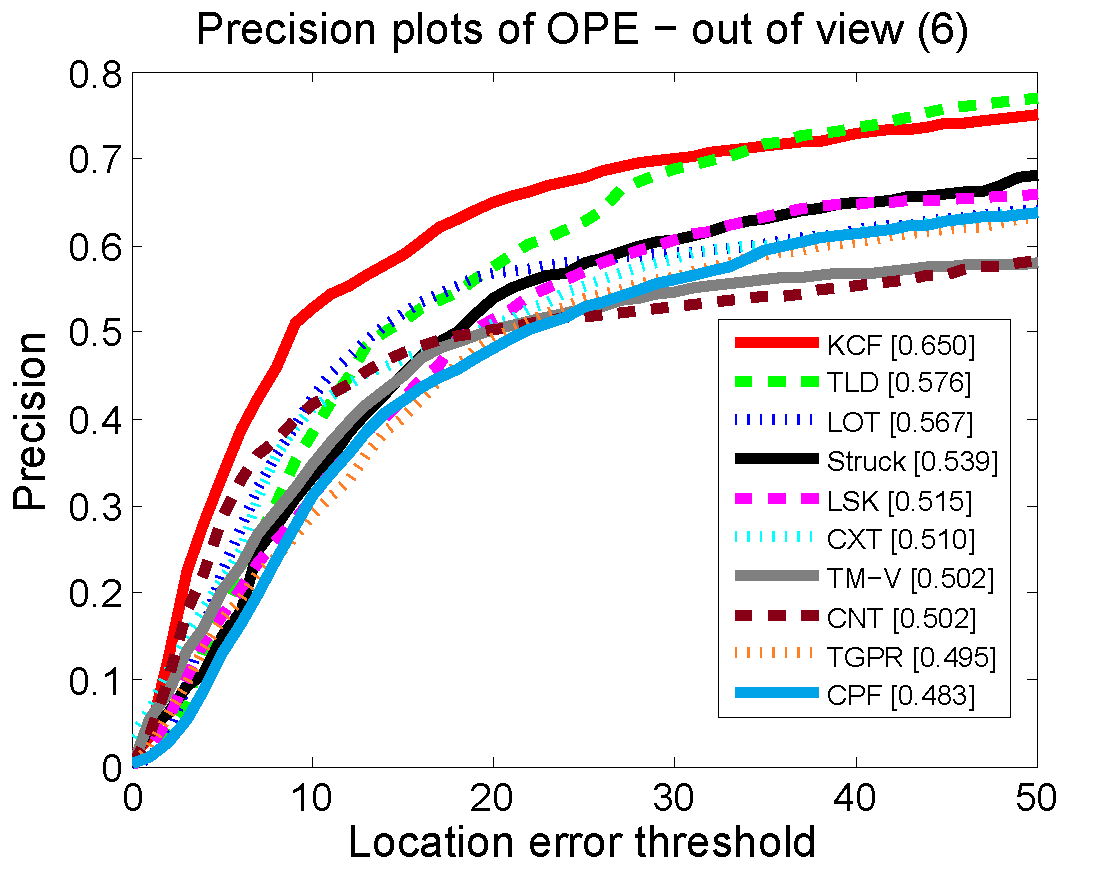}
\end{tabular}
\end{center}
   \caption{The precision plots of videos with different attributes. Best viewed on color display.}
\label{fig:attributePrecision}
\end{figure*}
Figure~\ref{fig:attributeSuccess}
shows the success plots of videos with different attributes, while Figure~\ref{fig:attributePrecision} shows the corresponding precision plots.
We note that the proposed CNT ranks
within top 3 on 7 out of 11 attributes in success plots, which outperforms DLT on all 11
attributes. In the precision plots, CNT ranks top 3 on 6 out of 11 attributes, and again outperforms DLT on all attributes.
Since the AUC score of the success plot is more accurate than the score at one position in the precision plot, as in~\cite{wu2013online}, in the following we mainly analyze the rankings based on the success plots.

On the videos with attribute of \textit{low resolution}, CNT ranks 1st among all evaluated trackers, which outperforms the 2nd counterpart MTT by $4.8\%$. The low resolution in the videos makes it difficult to extract effective hand-crafted features from the targets, thereby leading to undesirable results. On the contrary, CNT can extract dense useful information across the entire target region by convolution operators, and hence provides enough discriminative information to accurately separate the target from background.

For the videos with attributes such as \textit{in-plane rotation,
out-of-plane rotation, scale variation}, and \textit{occlusion}, CNT ranks
2nd on all evaluated algorithms with a narrow margin (about 1 percent) to the 1st trackers, such as KCF, TGPR, and SCM. All these methods employ the local image features as image representations. KCF utilizes the HOG features to describe the target and its local context region, and TGPR extracts the covariance descriptors from the local image patches as image representations. Furthermore, both CNT and SCM
employ local features extracted from the normalized local image
patches. CNT exploits the useful local features across the target via
filtering while SCM learns the local features from the target and
background with sparse representation. Furthermore, both CNT and SCM
utilize the target template from the first frame to handle drift
problem.
The results indicate that the local representation of the target
and the target template in the first frame have a positive effect on
handling the above-mentioned attributes.

On the videos with \textit{deformation} and \textit{background clutter} attributes,
CNT ranks 3rd which follows KCF
and TGPR. KCF exploits dense HOG descriptors with
predefined spatial structures to represent the target while TGPR explores similar spatial structures in which covariance descriptors are extracted, which encode them the local geometric layout information of the target, thereby rendering them capability to effectively handle deformation. CNT encodes the
geometric layout information to multiple simple cell feature maps (see
Figure~\ref{fig:simplecellfeature}), which are stacked together to a
global representation, thereby equipping it with tolerance to
deformation. Furthermore, CNT employs the useful background context information that is online updated and pooled in every frame,  and hence provides helpful information to accurately locate the target from background clutter.

On the videos with the \textit{illumination variation} and \textit{motion blur} attributes,
CNT ranks 4th while TGPR and KCF rank top 2.
All these methods take advantage of normalized local image
information, which is robust to illumination variation. Furthermore, when the target appearance changes greatly due to motion blur, the relatively unchanged background exploited by these methods can provide useful information to help localize the target.
Finally, for the videos with \textit{fast motion} attribute, CNT ranks 5th while the top 4 trackers are Struck, KCF, TGPR, and TLD.
CNT does not address fast motion well due to the simple dynamic model based
on stochastic search, and so do SCM and ASLA.
On the contrary, the trackers based on dense sampling (e.g., Struck, KCF, TGPR, and
TLD) perform much better than others in the subset of fast motion
due to their large search ranges.
The performance of our CNT can be further
improved with more complex dynamic models, by reducing the image resolution that equals to increasing the search range, or with more particles in
larger ranges.
The fast motion attribute also affects the performance of trackers on other attribute, such as the \textit{out of view} attribute. There are 6 videos with out of view attribute but 5 videos out of them also have the fast motion attribute. Thus, KCF and Struck which work very well on the fast motion attribute also perform favorably with the out of view attribute. Furthermore, Struck employs a budgeting mechanism that can maintain the useful target samples from the entire tracking sequences, thereby it can redetect the target when it reappears after out of view, and hence results in a favorable result. Meanwhile, CNT explores the stable target information from the first frame, which helps redetect the object.

\subsection{Qualitative Comparisons}
\subsubsection{Deformation}
\begin{figure*}[t]
\begin{center}
\begin{tabular}{c}
\includegraphics[width=0.9\linewidth]{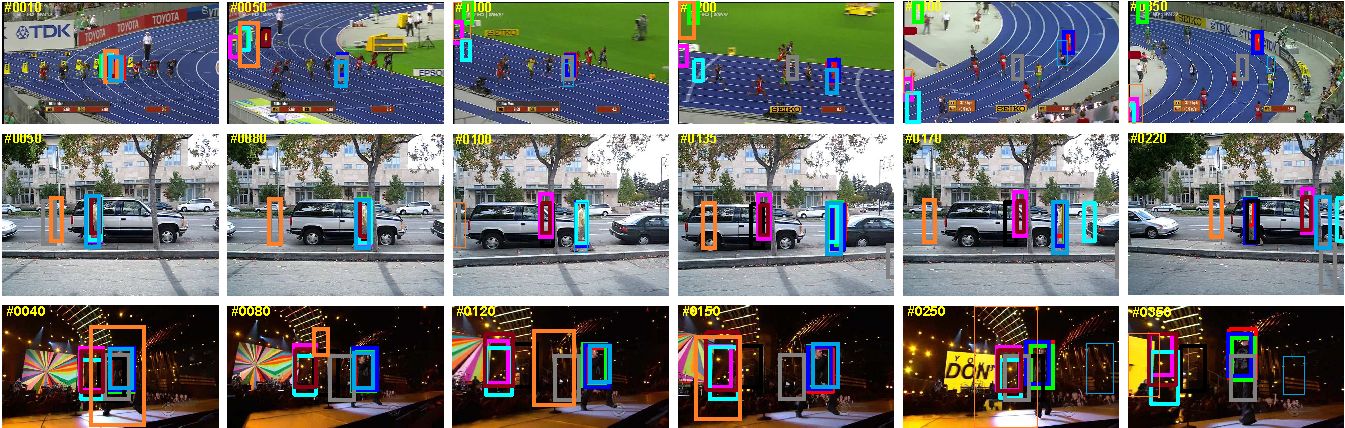}\\
\includegraphics[width=0.5\linewidth]{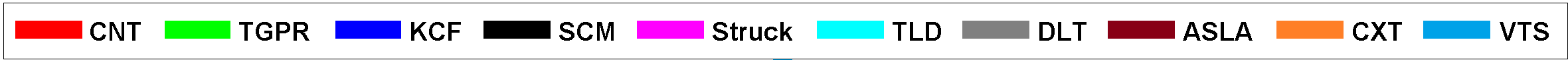}
\end{tabular}
\end{center}
\caption{Qualitative results of the 10 trackers over sequences \textit{bolt}, \textit{david3} and \textit{singer2}, in which the targets undergo severe deformation. Best viewed on color display.}
\label{fig:attributeDeformation}
\end{figure*}
\begin{figure*}[t]
\begin{center}
\begin{tabular}{c}
\includegraphics[width=0.9\linewidth]{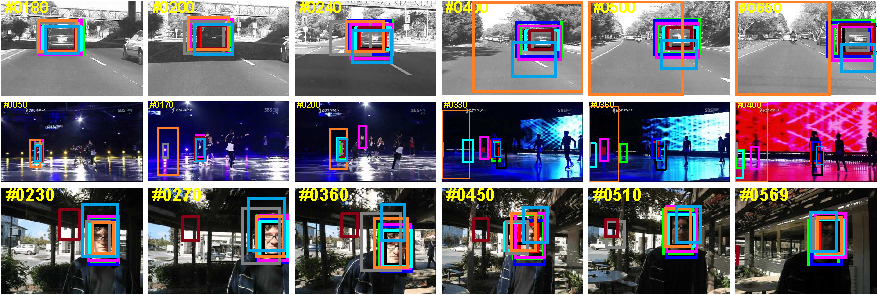}\\
\includegraphics[width=0.5\linewidth]{figs/legend.png}
\end{tabular}
\end{center}
\caption{Qualitative results of the 10 trackers over sequences \textit{car4}, \textit{skating1} and \textit{trellis}, in which the targets undergo severe illumination variations. Best viewed on color display.}
\label{fig:attributeIllumination}
\end{figure*}
Figure~\ref{fig:attributeDeformation} illustrates some screenshots of the tracking results in three challenging sequences where the target appearances undergo severe deformation. In the \textit{bolt} sequence, several objects appear in the screen with rapid appearance changes due to shape deformation and fast motion. Only the CNT and KCF algorithms can track the target stably. The TGPR, SCM, TLD, ASLA, CXT and VTS methods undergo severe drift at the beginning of the sequence (e.g.$\#10, \#100$). The DLT algorithm drifts to the background at frame $\#200$. The target in the $david3$ sequence suffers from significant appearance variations due to non-rigid body deformation. Furthermore, the target appearance changes drastically when the person walks behind the tree and when he turns back, thereby increasing difficulty to robustly tracking the target. The DLT and CXT algorithms lose tracking the target after frame $\#50$. The SCM, ALSA and VTS methods snap to some parts of background when the man walks behind the tree (e.g., $\#100$, $\#135$, $\#170$). The TLD algorithm loses the target when the man turns back at frame $\#135$. The Struck method snaps to the background when the person walks behind the tree again (e.g., $\#220$). Only the CNT, TGPR and KCF methods perform well at all frames. The target in the $singer2$ sequence undergoes both deformation and illumination variations. Only the CNT, TGPR and KCF algorithms perform well in the entire sequence. The proposed CNT handles deformation well because it employs the sparse local features with an adaptive threshold in (\ref{eq:sparseFeatureMap}) that can effectively filter out the varying parts in the appearance.
\subsubsection{Illumination Changes}
Figure~\ref{fig:attributeIllumination} shows some sampled results in three sequences in which the targets undergo severe illumination variations. In the \textit{car4} sequence, a moving vehicle passes beneath a bridge and under trees. Although the target undergoes drastic illumination variations at frames $\#180$, $\#200$, $\#240$,
the CNT method is able to track the object well. The DLT, CXT and VTS algorithms suffer from drift when the target undergoes a sudden illumination change at frame $\#240$. Furthermore, the target also undergoes obvious scale variations (e.g. $\#500$, $\#650$). Although the TGPR and KCF methods are able to successfully track the target, they cannot handle scale variations well (e.g., $\#500$, $\#650$). The target in the $skating1$ sequence undergoes rapid pose variations and drastic light changes (e.g., $\#170$, $\#360$, $\#400$). Only the CNT and KCF can persistently track the object from the beginning to the end. In the \textit{trellis} sequence, a person moves underneath a trellis with large illumination change and cast shadows while changing his pose, resulting in a significant variation in appearance. The DLT and ASLA methods drift away to background (e.g., $\#510$). The CNT, TLD and Struck methods are able to stably track the target, with much more accurately results than the TGPR, KCF and CXT methods that can persistently track the target. The CNT algorithm deals with illumination variations well because it extracts features via the normalized local filters with local brightness and contrast normalization.

\subsubsection{Scale Variations}
\begin{figure*}[t]
\begin{center}
\begin{tabular}{c}
\includegraphics[width=1\linewidth]{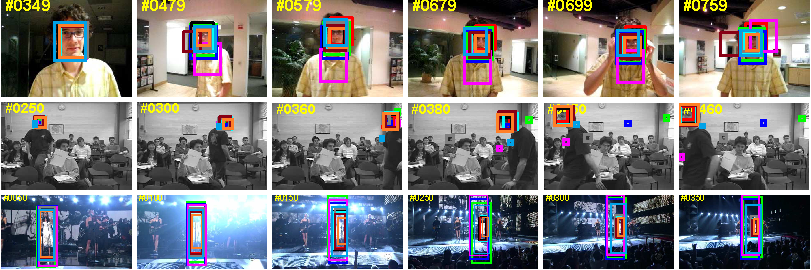}\\
\includegraphics[width=0.5\linewidth]{figs/legend.png}
\end{tabular}
\end{center}
\caption{Qualitative results of the 10 trackers over sequences \textit{david}, \textit{freeman3} and \textit{singer1}, in which the targets undergo scale variations. Best viewed on color display.}
\label{fig:attributeScale}
\end{figure*}
Figure~\ref{fig:attributeScale} demonstrates some results over three challenging sequences with targets undergoing significant scale variations. In the \textit{david} sequence, a person moves from a dark room to a bright area while his appearance changes much due to illumination variation, pose variation, and a large scale variation in the target relative to the camera. The ASLA and VTS algorithms drift away to background (e.g. $\#479$, $\#759$). The KCF and Struck methods do not track the target scale, resulting in smaller success rate of their results on the attribute of scale variation than the CNT method. In the \textit{freeman3} sequence, a person moves toward the camera, leading to a large scale variation in his face appearance. Furthermore, the appearance also changes much due to pose variation and low resolution, thereby increasing difficulty to accurately estimate its scale variation.  The TGPR, KCF, Struck, DLT and VTS methods suffer from severe drift (e.g., $\#380$, $\#450$, $\#460$). The CNT, SCM, TLD and CXT algorithms perform well. In the \textit{singer1} sequence, the target moves far away from the camera, resulting in a large scale variation. The TGPR, KCF, Struck and VTS methods cannot perform well while the CNT, SCM, ASLA and CXT algorithm achieve much better performance. The CNT handles scale variation well because its representation is built on scale-invariant complex cell features (See Figure~\ref{fig:complexcellftr}).

\subsubsection{Heavy Occlusion}
\begin{figure*}[t]
\begin{center}
\begin{tabular}{c}
\includegraphics[width=1\linewidth]{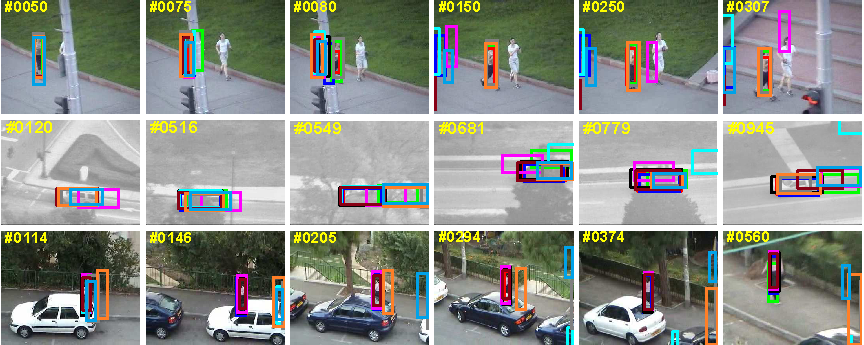}\\
\includegraphics[width=0.5\linewidth]{figs/legend.png}
\end{tabular}
\end{center}
\caption{Qualitative results of the 10 trackers over sequences \textit{jogging-1}, \textit{suv} and \textit{woman}, in which the targets undergo scale variations. Best viewed on color display.}
\label{fig:attributeOcclusion}
\end{figure*}
Figure~\ref{fig:attributeOcclusion} shows some sampled results of three sequences where the targets undergo heavy occlusion. In the \textit{jogging-1} sequence, a person is almost completely occluded by the lamppost (e.g., $\#75$, $\#80$). Only the CNT, TGPR, DLT and CXT algorithms are able to re-detect the object when the person reappears in the screen (e.g., $\#80$, $\#150$, $\#250$). In the \textit{suv} sequence, a vehicle undergoes heavy occlusions several times from dense tree branches (e.g., $\#516$, $\#549$, $\#681$, $\#799$), which makes it very challenging to accurately track the object. The TGPR, Struck, TLD, ASLA and VTS methods cannot perform well (e.g., $\#945$). In the \textit{woman} sequence, almost half body of a person is occluded by cars several times (e.g., $\#114$, $\#374$). The CNT, TGPR, KCF, SCM, Struck and ASLA algorithms achieve favorable results. All these methods employ the local features that are robust to occlusions.
\subsection{Analysis of CNT}
\begin{figure}[t]
\begin{center}
\begin{tabular}{c}
\includegraphics[width=0.5\linewidth]{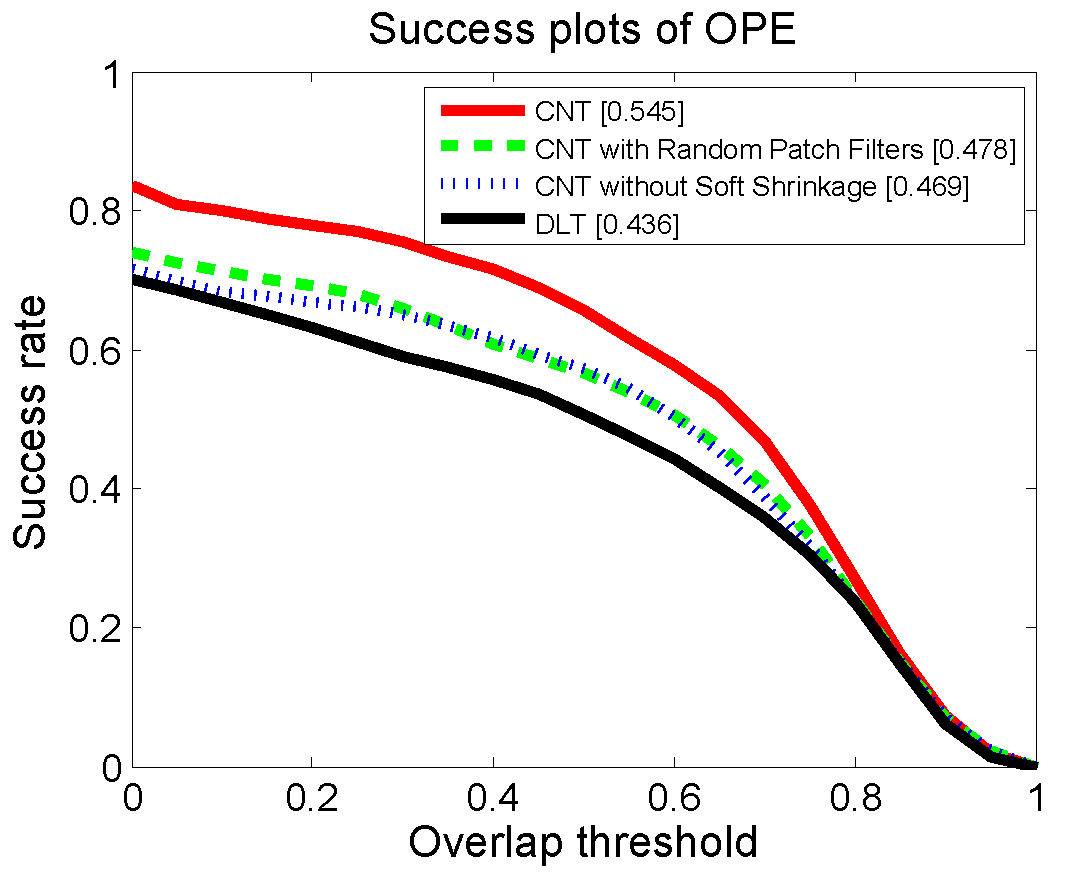}
\includegraphics[width=0.5\linewidth]{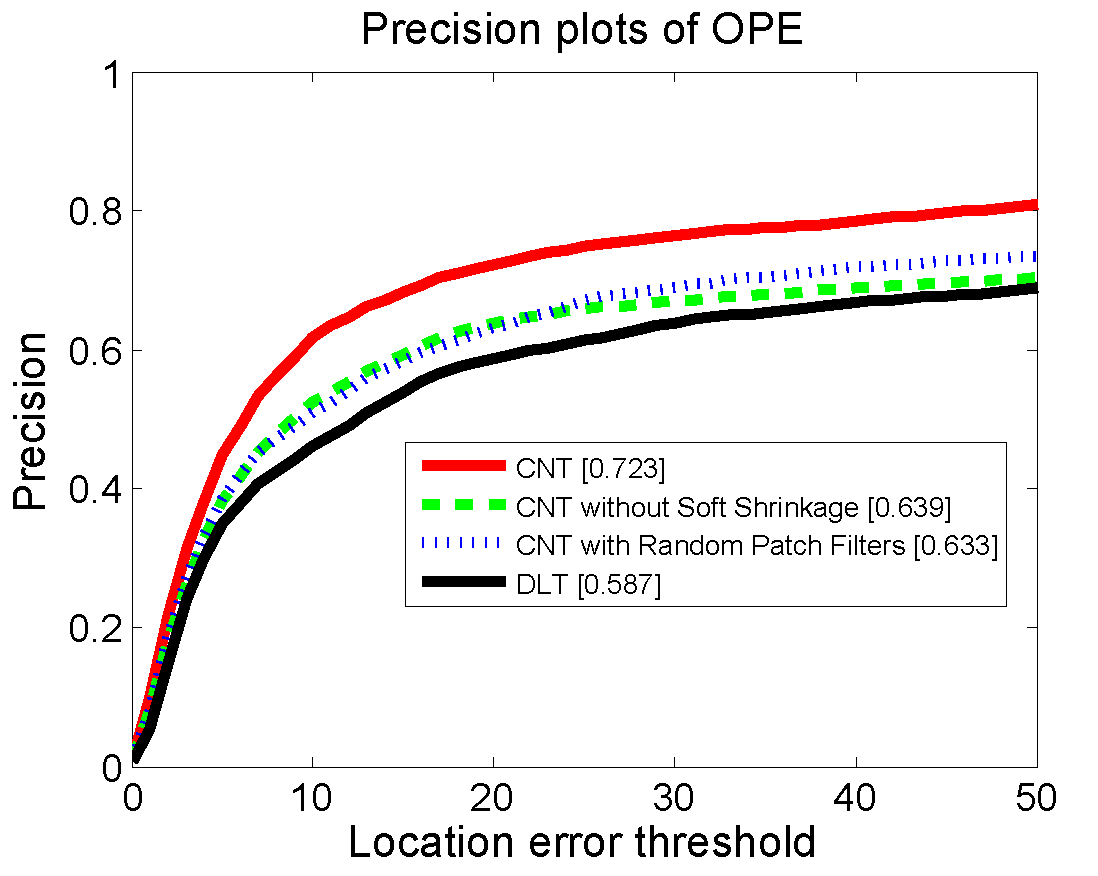}
\end{tabular}
\end{center}
\caption{The success plots and precision plots of OPE for CNT with different components. The DLT is taken as a baseline.}
\label{fig:discussion}
\end{figure}
To verify the effectiveness of some key components of CNT, we propose two variants of CNT: one is to utilize random patch filters to replace the filters learned by $k$-means algorithm in CNT, and the other is without the soft shrinkage component in CNT. Figure~\ref{fig:discussion} shows their quantitative results on the benchmark dataset. We can observe that with random patch filters, the AUC score of success rate reduces about $7\%$. Meanwhile, the CNT without soft shrinkage can only achieve AUC score 0.469, following the original CNT method 0.545 by a large margin. Notwithstanding, both variants perform better than the DLT method. We can conclude that both the filters and the soft shrinkage components play a key role in determining the performance of CNT.
\section{Concluding Remarks}
In this paper, we have proposed a simple two-layer feed-forward convolutional network that is powerful enough to produce an effective representation for robust tracking. The first layer is constructed by a set of simple cell feature maps defined by a bank of filters, in which each filter is a normalized patch extracted from the first frame with simple $k$-means algorithm, and then in the second layer, the simple cell feature maps are stacked to a complex cell feature map as the target representation, which encodes the local structural and geometric layout information of the target. A simple soft shrinkage strategy is employed to de-noise the target representation. A simple and effective online scheme is adopted to update the representation, which adapts to the target appearance variations during tracking. Extensive evaluation on a large benchmark dataset demonstrates the proposed tracking algorithm achieves favorable results against some state-of-the-art methods.

There are several possible directions to extend this work. First, some more effective selection strategies can be exploited to choose a set of more informative filters. Second, it is interesting to take into account a discriminative tracking framework, which employs the proposed two-layer convolutional network as a feature extractor.

\bibliographystyle{ieeetr}
\bibliography{egbib}
\end{document}